\pdfoutput=1

\documentclass[11pt]{article}

\usepackage[final]{acl}

\usepackage{times}
\usepackage{latexsym}

\usepackage[T1]{fontenc}

\usepackage[utf8]{inputenc}

\usepackage{microtype}

\usepackage{inconsolata}

\usepackage{hyperref}
\usepackage{graphicx}
\usepackage{amsmath}
\usepackage{amssymb}
\usepackage{listings}
\usepackage{tcolorbox}
\usepackage{cleveref}

\title{PsychoGAT: A Novel Psychological Measurement  Paradigm through Interactive Fiction Games with LLM Agents}

\author{Qisen Yang$^{1}$\thanks{Equal contribution, ordered alphabetically. Work was done during Zekun Wang's internship at Tsinghua University.}, Zekun Wang$^{2}$\footnotemark[1], Honghui Chen$^{3}$, Shenzhi Wang$^{1}$, Yifan Pu$^{1}$, Xin Gao$^{4}$,\\
   \bf Wenhao Huang$^{2}$, Shiji Song$^{1}$, Gao Huang$^{1}$\thanks{Corresponding author.} \\
  $^1$ Department of Automation, BNRist, Tsinghua University \quad
  $^2$01.AI \\ $^3$Medical Psychological Center, The Second Xiangya Hospital, Central South University \\ \& Medical Psychological Institute, Central South University \\ \& National Clinical Research Center for Mental Disorders \\ 
  $^4$ Carnegie Mellon University
}

\begin{document}
\maketitle

\begingroup
\renewcommand{\thefootnote}{}
\footnotetext{Emails: yangqs19@mails.tsinghua.edu.cn, noah.zenmoo re@gmail.com, choraschan@csu.edu.cn, \{wsz21, puyf23\}@ mails.tsinghua.edu.cn, xing1@andrew.cmu.edu, rubio8741@ gmail.com, \{shijis, gaohuang\}@tsinghua.edu.cn.}
\endgroup

\begin{abstract}
Psychological measurement is essential for mental health, self-understanding, and personal development. 
Traditional methods, such as self-report scales and psychologist interviews, often face challenges with engagement and accessibility.
While game-based and LLM-based tools have been explored to improve user interest and automate assessment, they struggle to balance engagement with generalizability.
In this work, we propose PsychoGAT (\textbf{Psycho}logical \textbf{G}ame \textbf{A}gen\textbf{T}s) to achieve a generic gamification of psychological assessment.
The main insight is that powerful LLMs can function both as adept psychologists and innovative game designers. 
By incorporating LLM agents into designated roles and carefully managing their interactions, PsychoGAT can transform any standardized scales into personalized and engaging interactive fiction games.
To validate the proposed method, we conduct psychometric evaluations to assess its effectiveness and employ human evaluators to examine the generated content across various psychological constructs, including depression, cognitive distortions, and personality traits. 
Results demonstrate that PsychoGAT serves as an effective assessment tool, achieving statistically significant excellence in psychometric metrics such as reliability, convergent validity, and discriminant validity. 
Moreover, human evaluations confirm PsychoGAT's enhancements in content coherence, interactivity, interest, immersion, and satisfaction.
\end{abstract}

\section{Introduction}
\label{sec:introduction}
Psychological assessment plays a pivotal role in mental health and self-understanding~\cite{meyer2001psychological}, such as depression measurement, cognitive distortion detection, and personality testing~\cite{osipow1966personality,hogan1991personality,bohane2017resilients,sharma2020computational,world2022world}.
Traditional methods for measuring these psychological constructs have mainly relied on standardized questionnaires (\textit{i.e.}, self-report scales)~\citep{urbina2014essentials} and psychologist-patient dialogues~\citep{tu2024towards}.
However, these approaches come with inherent limitations: 
(1) standardized scales can be monotonous, reducing engagement, particularly among younger individuals, leading to resistance~\citep{merry2012effectiveness};
(2) expert interviews require significant human resources and are scarce, often making psychological assessments inaccessible to those in need~\citep{Crisp2014-global-supply}. 

\begin{figure}[!t]
    \centering
    \includegraphics[width=.5\textwidth]{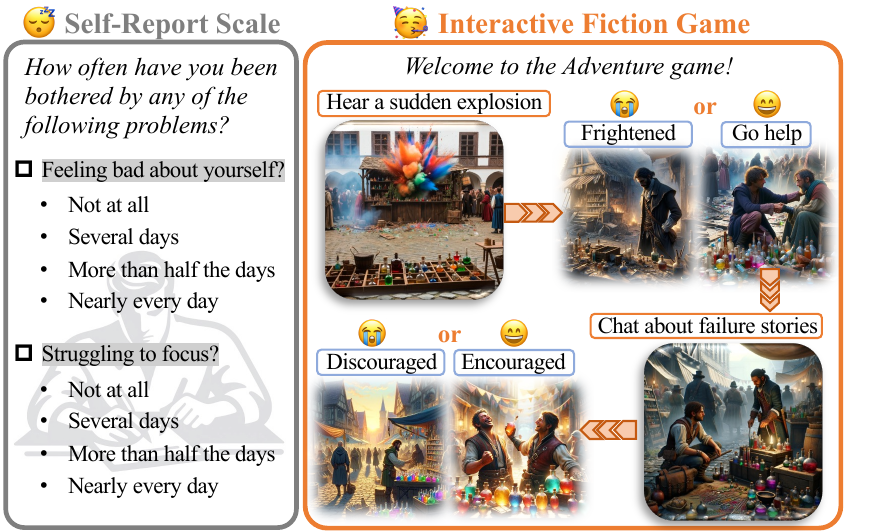}
    \caption{
    We propose PsychoGAT, a novel psychological assessment paradigm in the form of an interactive game instead of a traditional self-report scale. 
    }
    \label{fig:teaser}
    \vspace{-20pt}
\end{figure}

To address these challenges, there has been a shift towards automated, procedural methods for psychological assessments, including Computerized Adaptive Testing~\cite{meijer1999computerized} and rule-based electronic games specifically designed for measurements~\cite{jones1984video,song2020validation}. 
Besides, some studies have explored the use of Natural Language Processing (NLP) systems to support mental health~\citep{Empathy2,Empathy1, saha-etal-2022-shoulder,sentiment_analysis_assessment_tool}. 
However, these solutions are often limited by fragmented functionality and narrow applicability. 
Particularly, rule-based psychological games lack flexibility and struggle to engage users due to their inherent rigidity~\citep{volum-etal-2022-craft}.

\begin{figure*}[t!]
    \centering
    \includegraphics[width=\textwidth]{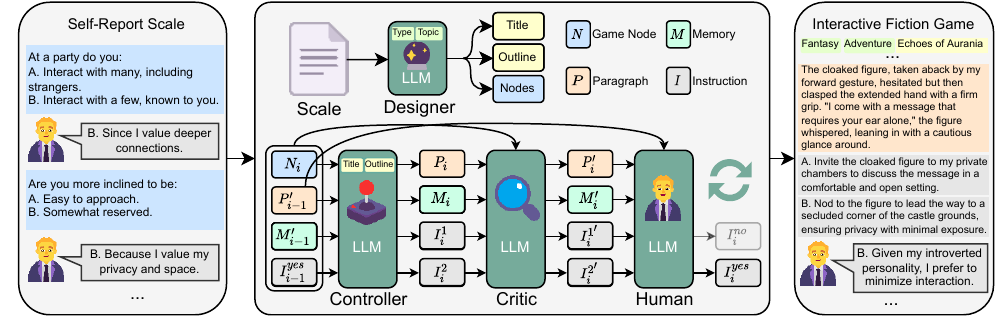}
    \caption{The multi-agent framework of PsychoGAT. The designer generates settings for the interactive fiction game based on a given self-report scale. The controller, critic, and a human participant (or human simulator) engage in a cyclical interaction to facilitate the assessment process. $I^{yes}$ represents the human-selected instruction.}
    \label{fig:framework}
    \vspace{-5pt}
\end{figure*}

Large language models (LLMs), known for their general-purpose capabilities, extensive knowledge, and reasoning abilities, offer a more flexible, knowledgeable, and engaging approach to psychological assessments~\citep{openai2023gpt4, he2023towards, llm-survey, palm-clinical-knowledge, wang2023interactive, sartori2023language, tu2024towards}. 
For example,  Diagnosis of Thought (DoT)~\citep{chen2023empowering_dot_prompting} utilizes LLMs to detect cognitive distortion via the chain of thought techniques~\citep{wei2022chainofthought}.
Another typical way for LLM-based psychological assessments is psychologist role-playing dialogue~\citep{tu2024towards, sartori2023language, wang2023rolellm}. 
Despite these advancements, current systems face limitations in generalizability across diverse psychological constructs, often due to their reliance on specific prior knowledge~\citep{chen2023empowering_dot_prompting,DBLP:conf/acl/0004RLWLMNA23}. 
Concurrently, psychologist role-playing dialogues often suffer from a lack of interestingness, leading to user disengagement~\citep{tu2024towards}.

In response to these issues, we propose a novel paradigm for interactive, immersive, and automatic psychological assessments, namely 
PsychoGAT (\textbf{Psycho}logical \textbf{G}ame \textbf{A}gen\textbf{T}s).
To the best of our knowledge, our method is the first one that incorporates LLM agents for generic and gamified psychological measurements. 
As illustrated in \Cref{fig:teaser}, our approach transforms traditional self-report scales for various psychological constructs into interactive fiction games, enabling players' interactions to serve as accurate indicators of their mental states. 
Specifically, as detailed in \Cref{fig:framework}, our framework comprises three main agents: (1) the game designer, responsible for creating the game's settings; (2) the game controller, tasked with generating interactive content; and (3) the critic, which refines the content to enhance the user experience.

In summary, our contributions~are:
\begin{itemize}
    \item We propose a novel paradigm for psychological measurements that shifts from traditional questionnaire-based methods to interactive, game-based assessments. The proposed paradigm enhances participant engagement and also ensures the assessments' psychometric reliability and validity.
    
    \item We develop an LLM-agent-based framework for the gamification of psychological assessments. This multi-agent framework leverages the capabilities of LLMs to facilitate dynamic interactions, allowing for its application across various assessment tasks and a wide range of game scenes.
    
    \item  Experimental results confirm the effectiveness and superiority of our approach, evidenced by competitive psychometric scores and significant improvements in user-experience-related metrics in human evaluations.
\end{itemize}

\section{PsychoGAT}
\label{sec: method}
As in Figure \ref{fig:framework}, our framework comprises three agents responsible for the game generation, in addition to a separate agent designed for simulating human interactions. The subsequent section will commence with an overview of the workflow (\S\ref{sec:workflow}), followed by detailed descriptions of each agent involved: the game designer (\S\ref{sec:designer}), the game controller (\S\ref{sec:controller}), the critic (\S\ref{sec:critic}), and the human simulator with a psychometric evaluator (\S\ref{sec:simulator}).
Detailed prompt templates are in Appendix~\ref{appx:prompts}.

\subsection{Overall Workflow}\label{sec:workflow}
Given a psychological construct, such as depression, we begin by selecting an associated self-report scale through retrieval or manual selection. 
This scale serves as external psychological knowledge guiding the creation of the interactive fiction game.
Additionally, players can choose both the game's type and topic for personalized experiences.

Subsequently, the \textbf{game designer} is tasked with crafting a game title, an outline, and corresponding game nodes ($N$). 
These nodes are composed of multiple, redesigned scale items, 
organized in alignment with the game's narrative and storyline. The title, outline, and list of nodes collectively serve as the conditions for subsequent game generation iterations. 
Then, based on the redesigned scale items, the \textbf{game controller} and \textbf{critic} agents engage in a recursive interaction. 
This process involves the generation and refinement of game paragraphs ($P$), the memories ($M$) to track previous game plots, and the instructions ($I$) for future game development. 
Through several iterations of this generative and refining process, the finely crafted instructions are presented to the player for selection, alongside recent game paragraphs and memories. 
In the whole process, a hard-coded \textbf{psychometric evaluator} collects the player's decisions at each iteration to calculate the assessment score (detailed in \S\ref{par:evaluator}).

\subsection{Game Designer}\label{sec:designer}

The designer creates a skeleton for the first-person interactive fiction game about the chosen type and topic, incorporating storylines that facilitate the detection of the player's psychological profile.
Inputs for this agent include the game type, game topic, and a traditional psychological assessment scale.
For example, the participant requires a fantasy game about adventure for the personality test, as illustrated in \Cref{fig:framework}.

We employ a chain-of-thought prompting technique~\citep{wei2022chainofthought} to enhance the game design. 
Specifically, the designer is prompted to first generate the game's title and the thoughts on the story's design—particularly focusing on how it might unveil aspects of the player's psychological profile—followed by the creation of a comprehensive game outline and the redesign of the scale to fit the game narrative. 
The redesigned scale is structured to mirror the game's outline, ensuring that the sequence of questions corresponds with the outline of the narrative. 
Each question is paired with two options, each linked to a binary score indicating the absence or presence of the psychological construct being detected.
This binary scoring system is designed to assess the player's psychological tendencies through their in-game decisions. 

\subsection{Game Controller}\label{sec:controller}

The controller's objective is to sequentially instantiate~scale items, facilitating the interaction between the player and the text-based game environment. 
Denote the controller agent and human player as $\pi_c$ and $\pi_p$. The process can be formulated as:

\vspace{-8pt}
{\small
\begin{equation}
\label{eq: game_designer_agent}
    \begin{aligned}
        (P_{i}, M_{i}, I^{1}_{i}, I^{2}_{i}) = \pi_c(N_{i}, P^{'}_{i-1}, M^{'}_{i-1}, I^{yes}_{i-1}; T, O),
    \end{aligned}
\end{equation}
}
where $N_{i}$ is the current game node (\textit{i.e.}, the redesigned scale item), $P^{'}_{i-1}$ is the previous paragraph after critic refinement, $M^{'}_{i-1}$ is the previous memory of game progress after critic refinement, $I^{1}_{i}$ and $I^{2}_{i}$ are the current instructions for the player to choose, $T$ is the game title, and $O$ is the game outline. $I^{yes}_{i-1} = \pi_p(P^{'}_{i-1}, I^{1'}_{i-1}, I^{2'}_{i-1})$ is the player selected instruction, where $I^{1'}_{i}$ and $I^{2'}_{i}$ are the corresponding refined instructions by the critic agent.

In the initial iteration, the controller generates the game's introduction and background. 
Subsequent interactions between the player, the critic, and the controller occur in a loop to instantiate each scale item in order. 
As the game progresses, the growing history becomes burdensome for the LLM to process efficiently.
To mitigate this issue, a summarization mechanism is employed to maintain coherent narrative memory~\citep{zhou2023recurrentgpt}.
It necessitates the model to justify exclusions of certain elements from prior memory and the inclusion of new information before updating the memory.

\begin{figure*}[t]
    \centering
    \includegraphics[width=.99\textwidth]{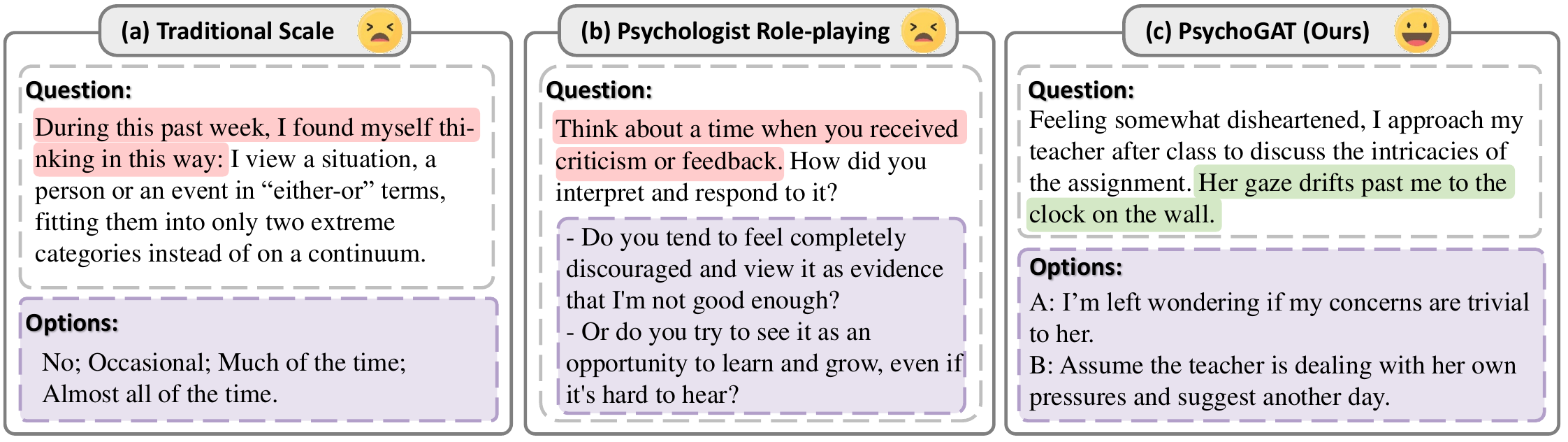}
    \caption{Comparison of assessment paradigms. Traditional scales and psychologist rolep-playing interviews~emphasize the recall of life events and self-reported feelings or thoughts. In contrast, PsychoGAT introduces an~interactive fiction game environment where participants make decisions as the protagonist and craft their personal story.}
    \label{fig:baseline_comparison_example}
    \vspace{-5pt}
\end{figure*}

\subsection{Critic}\label{sec:critic}

The critic aims to improve the overall quality of content generated by the controller.
Following the notations in Equation~\eqref{eq: game_designer_agent} and denote the critic agent as $\pi_{cri}$, the critic interactions can be formulated as 

{\small
\begin{equation}
    \begin{aligned}
        (P_{i}^{'}, M_{i}^{'}, I^{1'}_{i}, I^{2'}_{i}) = \pi_{cri}(P_{i}, M_{i}, I^{1}_{i}, I^{2}_{i}).
    \end{aligned}
\end{equation}
}

The controller and critic interact iteratively at each timestep until the content meets approval or a predefined maximum number of iterations is reached. 
Specifically, the critic aims to mitigate three main issues of the controller's outputs:

\textbf{The coherence issue} is inevitable due to numerous iterations with extensive textual content in our fiction game.
Despite efforts to enhance coherence using the memory mechanism discussed in \S\ref{sec:controller}, the controller only provides coarse-grained control and lacks detailed analysis for each iteration. 
Therefore, the critic is tasked with improving the alignment of generated paragraphs with the memory, preceding text, and player-chosen instructions.

\textbf{The bias issue\label{bias-issue}} arises from the tendency of generated paragraphs in successive iterations to be influenced and biased by previous human selections.
Specifically, when a player selects a certain option at one iteration, subsequent generated paragraphs may exhibit a default preference, leading to uniformity in the continuation instructions for interactions.
To mitigate this issue, we instruct the critic to assess whether the generated paragraph refrains from indicating a specific choice to the player and ensures that the provided instructions are distinct.

\textbf{The omission issue} refers to the controller's propensity to miss crucial details in extensive prompts, such as the failure to use a first-person perspective, which is essential for achieving deep player immersion in interactive fiction games. 
To solve this, we re-emphasize certain frequently overlooked requirements in the critic's prompt, such as maintaining a first-person narrative perspective and incorporating common expressions.

\subsection{Human Simulator and Psychometric Evaluator}\label{sec:simulator}

\paragraph{Human Simulator.} 
To facilitate automated operation and evaluation of our framework, and drawing upon the validated psychological characteristics and annotation capabilities of GPT-4~\citep{psychobench, gpt4-better-annotator-1, chatgpt-better-annotators, llm-as-a-judge, wang2023rolellm, safdari2023personality, ultrachat}, we employ GPT-4 to simulate human interactions in selecting instructions. 
At each iteration, GPT-4 is prompted with descriptions and examples illustrating the presence or absence of the specified psychological construct. 
The instruction selection is based on the narrative content of the preceding and current paragraphs, alongside the maintained memory of the game's progression. 
This process also incorporates the chain-of-thought technique~\citep{wei2022chainofthought}.

\paragraph{Psychometric Evaluator.\label{par:evaluator}} 
The instruction selections made by the human simulator at each iteration serve as indicators of its psychological tendencies. 
Given that each instruction choice corresponds to a particular option within a scale item, complete with an associated score, we deploy a psychometric evaluator to aggregate these scores. 
Consequently, by tallying the accumulated scores, we derive a comprehensive score that encapsulates the psychological profile of the human simulator.

\section{Experiments}
\label{sec: experiment}
This Section introduces the experimental tasks, baseline methods, and evaluation protocol in \S \ref{sec: experi_setup}. Then, the evaluation metrics are detailed in \S \ref{sec: experi_indices}. Next, we assess PsychoGAT's effectiveness in psychological measurements and compare various LLM-based assessment methods in \S \ref{sec: experi_Validation_and_Comparison}. Furthermore, ablation studies on PsychoGAT agents and game scene robustness are presented in \S \ref{sec: Ablation_and_case_Studies}. 
More experimental discussions are in Appendix~\ref{more_experimental_discussions}.

\subsection{Experimental Setup}
\label{sec: experi_setup}
\paragraph{Psychological Assessment Tasks.} We investigate various psychological assessment tasks including personality testing, depression measurement, and cognitive distortion detection. Specifically, experiments are conducted on the extroversion subscale of the Myers–Briggs Type Indicator (MBTI) personality test~\cite{boyle1995myers}, the patient health questionnaire (PHQ-9) for depression measuring~\cite{kroenke2001phq}, and the cognitive distortion detection of three common thinking traps (\textit{i.e.,} A. all-or-nothing, B. mind reading, and C. should statement)~\cite{kostoglou2016cognitive,DBLP:conf/acl/0004RLWLMNA23}.
Detailed scales with the explanation of each psychological construct are presented in \Cref{appx:scales}. 

\paragraph{Baseline Methods.} To measure the psychological effectiveness of PsychoGAT, we first compare it with the traditional scale (T-Scale), \textit{i.e.}, standardized self-report assessment~\cite{gault1907history,kjell2023beyond}. Furthermore, to evaluate the superiority of PsychoGAT, we compare it with other LLM-based psychological assessment tools. 
These include automatic scale generation (Auto-Scale), psychologist role-playing interview (Psycho-Interview), and enhanced interview specifically designed for cognitive distortion detection with Diagnosis of Thought Prompting~\citep{chen2023empowering_dot_prompting} (DoT-Interview). 
The comparisons across the three paradigms—scales, role-playing interviews, and interactive fiction games—are depicted in \Cref{fig:baseline_comparison_example}.
Detailed implementations of baseline methods are presented in \Cref{appx:baselines}.

For fair comparison, all LLM-based methods above use GPT-4~\cite{achiam2023gpt} at a temperature setting of 0.5, without additional training or fine-tuning.
In PsychoGAT, the critic-controller interaction is limited to a maximum of 3 iterations, and the player interaction is capped at 10 iterations.

\paragraph{Evaluation Protocol.} 
LLMs can exhibit stable personalities \cite{safdari2023personality,cui2023machine}, express nuanced emotions \cite{tavast2022language}, mimic human cognition \cite{schramowski2022large,binz2023using}, and facilitate various social science simulations \cite{aher2023using}.
Therefore, our study employs GPT-4 to simulate participants in psychological assessments to avoid potential ethical risks. 
Then, these simulated assessments are evaluated for their psychometric effectiveness and the user experience. 
The evaluation involves two aspects: automatic evaluations using psychometric indicators and human evaluations through content-related metrics. 

\subsection{Evaluation Metrics}
\label{sec: experi_indices}

\paragraph{Psychometric Evaluation.}
In psychometrics, the central criteria of scientific assessments are \textit{reliability} and \textit{construct validity}, which underscore the ability of an assessment tool to reliably and accurately capture the essence of the latent construct it aims to measure~\cite{safdari2023personality}.

\textbf{Reliability} refers to the inner consistency and dependability of the assessment. In this work, we employ two reliability metrics: the commonly used Cronbach's Alpha ($\alpha$)~\cite{cronbach1951coefficient,zinbarg2005cronbach} and the less biased Guttman's Lambda 6 ($\lambda_6$)~\cite{guttman1945basis}, formulated as follows.

{\small
\begin{equation}
\label{alpha equation}
    \begin{aligned}
        \alpha = \frac{k}{k-1}\left(1-\frac{\sum_{i=1}^k \sigma_y^2}{\sigma^2_x}\right),\ \lambda_6 = 1-\frac{\sum_{i=1}^k e_i^2}{\sigma^2_x},
    \end{aligned}
\end{equation}
}
where $k$ is the number of items (\textit{i.e.}, questions) on the test, $\sigma_y^2$ is the variance associated with each item $i$, $\sigma_x^2$ is the variance of the total scores, and $e_i^2$ is the variance of the errors of estimate of item $i$ from its linear multiple regression on the remaining $k-1$ items.

\textbf{Construct validity} can be assessed via \textit{convergent validity}, and \textit{discriminant validity}~\cite{clark2019constructing}.
Convergent validity assesses whether a test correlates to established indicators of its target construct, while discriminant validity shows how sufficiently uncorrelated a test is to indicators of unrelated constructs.
The convergent and discriminant validity are evaluated using Campbell's classic framework~\cite{campbell1959convergent}, operationalized as Pearson product-moment correlations:

{\small
\begin{equation}
\label{pearson coefficient}
    \begin{aligned}
        r_{xy} = \frac{\sum_{i=1}^n(x_i-\Bar{x})(y_i-\Bar{y})}{\sqrt{\sum_{i=1}^n(x_i-\Bar{x})^2}\sqrt{\sum_{i=1}^n(y_i-\Bar{y})^2}},
    \end{aligned}
\end{equation}
}
where $n$ is the sample size, $x_i, y_i$ are a pair of data points from samples, $\Bar{x}, \Bar{y}$ are the mean scores.

\begin{table*}[t]
\centering
\small
\begin{tabular}{lccccccccc}
\hline
\textbf{Task} & \multicolumn{2}{c}{\textbf{Reliability ($\alpha$)}} & \multicolumn{2}{c}{\textbf{Reliability ($\lambda_6$)}}  & \multicolumn{2}{c}{\textbf{Convrgent Validity}}  & \multicolumn{2}{c}{\textbf{Discriminant Validtity}} & \textbf{Overall}\\
\hline
Personality & 0.97 & + + + &  0.98 &+ + + & 0.97 &+ & -0.59  &+ & + + + \\
Depression & 0.77 &+ &  0.84 &+ + & 0.85   &+  & -0.07  &+  & + \\
Cognitive Distortion-A & 0.92 &+ + + & 0.93 &+ + + & 0.97  &+  & -0.44  &+ 
 & + + + \\
Cognitive Distortion-B	& 0.92 &+ + + & 0.95 &+ + + & 0.97  &+  & 0.25  &+  & + + + \\
Cognitive Distortion-C	& 0.88 &+ + &  0.91 &+ + + & 0.93  &+  & -0.18   &+  
 & + + \\
\hline
\end{tabular}
\caption{\label{main_results_simulation}
The psychometric evaluations of PsychoGAT. Results show an overall significant reliability and construct validity (Statistically, + acceptable, ++ good, +++ excellent), verifying PsychoGAT as a scientific measurement~tool. 
}
\end{table*}

\begin{figure*}[!t]
    \centering
    \includegraphics[width=.98\textwidth]{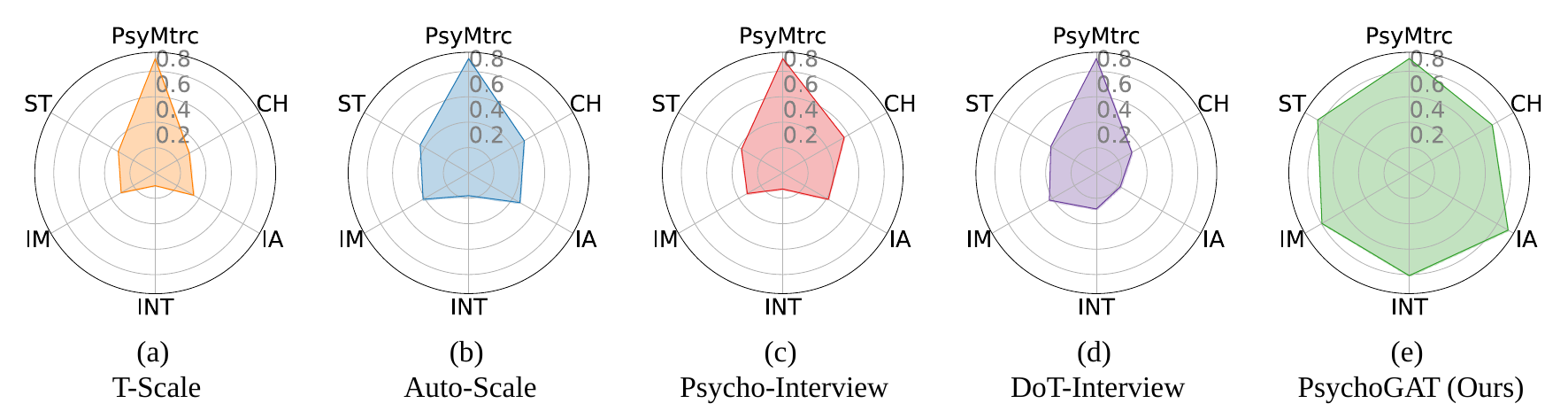}
    \vspace{-5pt}
    \caption{Comparisons among various assessment methods through automatic and human evaluations. All five methods are qualified for psychometric effectiveness, and PsychoGAT brings all-around user experience~improvement. PsyMtrc= Psychometric, CH= Coherence, IA= Interactivity, INT= Interest, IM= Immersion, and ST= Satisfaction.}
    \label{fig:baseline_comparison}
    \vspace{-5pt}
\end{figure*}

In our experiments, we designate a given reliability metric (RM; \textit{i.e.}, $\alpha$, $\lambda_6$) $<$ 0.50 as unacceptable, 0.50 $\leq$ RM $<$ 0.60 as poor, 0.60 $\leq$ RM $<$ 0.70 as questionable, 0.70 $\leq$ RM $<$ 0.80 as acceptable (denoted as +), 0.80 $\leq$ RM $<$ 0.90 as good (+ +), and RM $\geq$ 0.90 as excellent (+ + +).
We use cut-offs~\cite{evans1996straightforward} for considering correlations as weak, moderate, strong, and very strong (\textit{i.e.} $\lvert r \rvert$ $<$ 0.40, 0.40 $\leq$ $\lvert r \rvert$ $<$ 0.60; 0.60 $\leq$ $\lvert r \rvert$ $<$ 0.80; 0.80 $\leq$ $\lvert r \rvert$; respectively).
Reliability validation requires RM $\geq$ 0.70, convergent validity requires $\lvert r \rvert$ $\geq$ 0.60 (denoted as +), and discriminant validity requires $\lvert r \rvert$ $\textless$ 0.60 (denoted as +).

\paragraph{Content Evaluation.}
We consider five dimensions for human evaluators to assess the generated content~\cite{scenecraft,ryan2015narrative,nacke2011towards,jennett2008measuring}: (\romannumeral1) \textit{Coherence (CH)}, assessing the logical consistency of the narrative; (\romannumeral2) \textit{Interactivity (IA)}, gauging how responsive and unbiased\footnote{Please refer to the bias issue in \S\ref{bias-issue}.} the content is to participants' interactions; (\romannumeral3) \textit{Interest (INT)}, indicating the appeal of the generated interactive fiction games; (\romannumeral4) \textit{Immersion (IM)}, measuring the extent of the participant's engagement with the narrative world;  (\romannumeral5) \textit{Satisfaction (ST)}, evaluating the overall user experience.

\subsection{Main Experiments}
\label{sec: experi_Validation_and_Comparison}

We first assess the psychometric effectiveness of PsychoGAT and then compare it with other LLM-based assessment methods. For each assessment task and for each method, experiments are conducted over 20 measurement samples from the human simulator, comprising an equal distribution of negative and positive instances of the assessed construct. 
Particularly for PsychoGAT, there are 10 common [type; topic] pairs during game generation, including [Fantasy; Adventure, Magic], [Romance; Love, Marriage], [Science Fiction; Space Exploration, Time Travel], [Slice of Life; Family, School], [Horror; Haunted House, Paranormal Investigation].
Furthermore, we recruit real human participants instead of the human simulator to ensure a robust assessment of our method. 

\paragraph{Psychometric Effectiveness of PsychoGAT.} 
Automatic evaluations are conducted to investigate the psychometric effectiveness of PsychoGAT.
The convergent validity is evaluated through the comparison of average correlations between PsychoGAT scores and those from established scales targeting the same construct (e.g., depression measures). Conversely, discriminant validity was assessed by examining the average correlations between PsychoGAT scores and those from scales measuring different constructs (i.e., learning style assessments).
As shown in \Cref{main_results_simulation}, PsychoGAT exhibits both high reliability and robust construct validity, including convergent and discriminant validity. 
These observations suggest that PsychoGAT constitutes a strongly reliable and accurate tool for psychological assessment.

\begin{figure}[h]
    \centering
    \includegraphics[width=.45\textwidth]{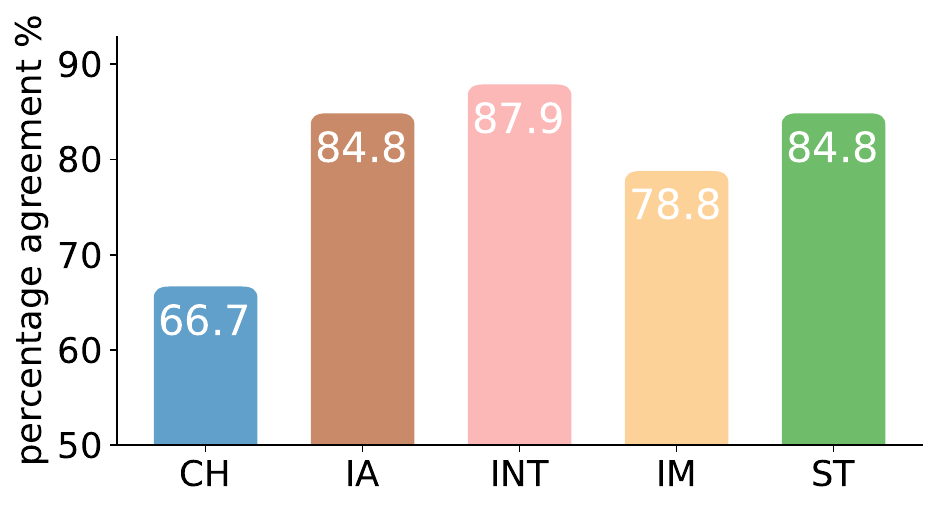}
    \vspace{-15pt}
    \caption{Percentage agreements of PsychoGAT's superiority across five human evaluation metrics, as defined in \Cref{fig:baseline_comparison} and \S\ref{sec: experi_indices}.}
    \label{fig:huamn_baseline_agree}
    \vspace{-17pt}
\end{figure}

\paragraph{Comparative Experiments.}
The four LLM-based assessment methods are first verified for psychometric effectiveness.
Subsequently, 33 evaluators with a basic knowledge of psychological assessments are recruited to evaluate these approaches.
We randomly select 15 pieces of assessment content from the participant simulation results on the cognitive distortion (all-or-nothing thinking) detection task. 
Evaluators are tasked with rating this content across five dimensions: coherence, interactivity, interest, immersion, and satisfaction, on a scale from 1 to 5.
The average scores are normalized to a range of [0.1, 0.9].
Results underscore the comprehensive superiority of PsychoGAT in enhancing the user experience, as illustrated in \Cref{fig:baseline_comparison}.
Additionally, the reliability of human evaluations is confirmed by the percentage agreements on PsychoGAT's superiority across the five dimensions, as shown in \Cref{fig:huamn_baseline_agree}.

\begin{figure}[h]
    \centering
    \includegraphics[width=.38\textwidth]{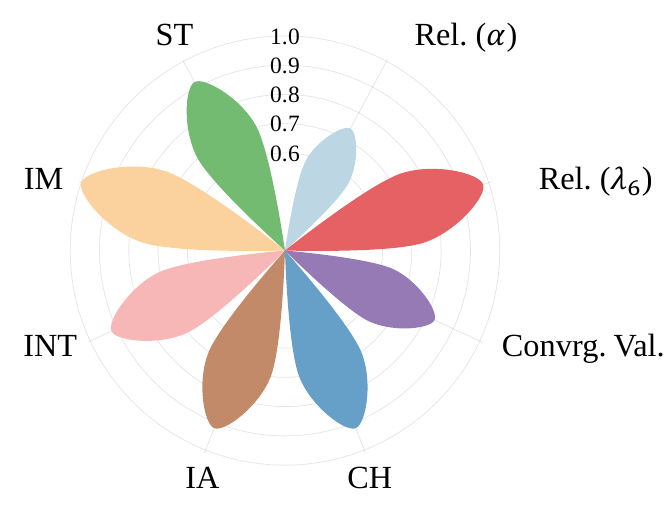}
    \vspace{-5pt}
    \caption{Human participant experiments on psychometric effectiveness (Reliability $\alpha$, $\lambda_6$, and convergent validity) and content evaluations (five human evaluation metrics as defined in \S\ref{sec: experi_indices}).}
    \label{fig:human_participant}
    \vspace{-5pt}
\end{figure}

\paragraph{Human Participant Experiments.}
Despite existing literature supporting the simulation validity of GPT-4 on various psychological traits, we conduct human participant experiments to further evaluate the proposed method. 
The traditional scale and our PsychoGAT are compared on the personality assessment task. 
12 participants aged 20 to 30, all proficient in English, have completed both the self-report scale and PsychoGAT, followed by a user experience questionnaire.
As shown in \Cref{fig:human_participant}, PsychoGAT demonstrates significant reliability and construct validity, consistent with our simulation evaluations presented in \Cref{main_results_simulation}.
Additionally, we compare user experience feedback for both the traditional and PsychoGAT assessment processes. 
The results, depicted in \Cref{fig:human_participant}, show a majority of participants indicating a preference for PsychoGAT across various metrics, aligning with the results in \Cref{fig:huamn_baseline_agree}.
More experimental details are presented in Appendix~\ref{sec: Details of Human Participant Experiments}.

\subsection{Ablation and Case Studies}
\label{sec: Ablation_and_case_Studies}

\textbf{Ablation on Game Scenes.} PsychoGAT incorporates a variety of game types and topics to enhance user engagement. 
To examine the psychometric robustness across different game scenes, our first ablation study focuses on analyzing the psychometric effectiveness of slice-of-life (SoL), science fiction (Sci-Fi), and horror games produced by PsychoGAT.
In \Cref{ablation_type}, each game type independently produces 20 simulation assessment samples on the personality test task.
Compared to the overall performance with diverse game types in \Cref{main_results_simulation}, assessments focused on a single type demonstrate equally excellent reliability and construct validity, which underscores the robustness of our method in providing reliable psychological measurements across various thematic contexts.

\begin{table}[h]
\small
\centering
\begin{tabular}{lccc}
\hline
\textbf{Scene} & \textbf{Reliability}  & \textbf{Convrg. Val.}  & \textbf{Disc. Val.} \\
\hline
SoL & 0.99 + + + &  0.99 + & -0.56 +  \\
Sci-Fi &  0.99 + + +  &  0.99 + & -0.58 +     \\
Horror & 0.98 + + + & 0.99 + & -0.59 +  
 \\
Overall	& 0.98 + + + & 0.97 +  & -0.59 +  \\
\hline
\end{tabular}
\caption{\label{ablation_type}
The ablation study to verify PsychoGAT's psychometric robustness among different game scenes on the personality assessment task. Please refer to \Cref{main_results_simulation} and \S\ref{sec: experi_indices} for the psychometric notation ``+''.}
\vspace{-10pt}
\end{table}

\begin{figure}[h]
    \centering
    \includegraphics[width=.48\textwidth]{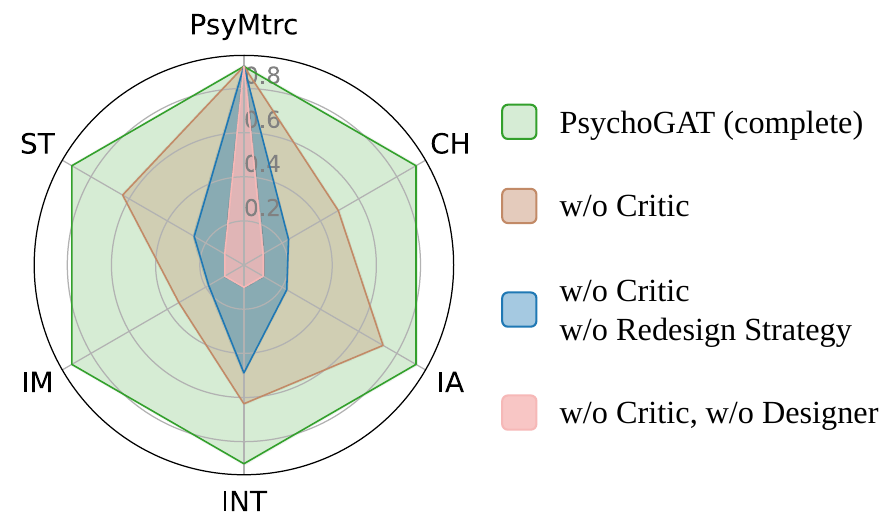}
    \vspace{-20pt}
    \caption{The ablation study to assess the effectiveness of PsychoGAT's three agents (\textit{i.e.}, the designer, controller, and critic) and the scale redesign strategy.}
    \label{fig:ablation}
    \vspace{-10pt}
\end{figure}

\paragraph{Ablation on Agents.} PsychoGAT comprises three agents (namely, the game controller, game designer, and critic) alongside a scale redesign strategy implemented by the game designer agent. 
To assess the effectiveness of these agents, we conduct ablation studies to evaluate their impact on the psychometric performance and user experience in the personality test task. 
Specifically, we collect 20 samples for each ablation study by participant simulations. For human evaluations, we enlist the expertise of 2 LLM researchers and 1 clinical psychology researcher to evaluate five content evaluation metrics, as detailed in \S \ref{sec: experi_indices}. 
The results, depicted in \Cref{fig:ablation}, indicate that these agents exert a minimal effect on psychometric effectiveness. 
It suggests that GPT-4 possesses inherent capabilities to ensure reliable measurements across various paradigms. 
Conversely, each agent significantly influences the quality of the generated games, underscoring their collective importance.

\section{Related Work}
\label{sec:related_work}
\paragraph{Gamified Assessments in Psychology.} 
Game-based psychological assessments not only achieve high accuracy but also increase engagement and enjoyment for participants~\cite{kim2016applying,turan2018game,derosier2019hall}. However, as an emerging interdisciplinary research area, gamification in psychological assessments are still in their initial stages. Researchers have crafted various online games to assess traits such as aggression~\cite{mccreery2019can}, identify and evaluate reading challenges in young students~\cite{hautala2020identification}, and gauge cognitive functions~\cite{song2020validation}. Yet, these studies primarily rely on the manual designs of psychologists and game designers, leading to a high investment of time, financial resources, and human effort.

\paragraph{LLM-assisted Psychology.} 
The integration of artificial intelligence, especially via LLMs, has opened up abundant new avenues for psychological research across educational, social, cultural, clinical, and counseling domains \cite{sun2021psyqa,ke2024exploring,sartori2023language,he2023towards,binz2023turning,park2023artificial,ziems2023can}.
LLMs exhibit stable personalities \cite{safdari2023personality,cui2023machine}, express nuanced emotions \cite{tavast2022language}, mimic human cognition \cite{schramowski2022large,binz2023using}, and facilitate various social science simulations \cite{aher2023using}.
In psychological assessments, LLMs introduce innovative approaches, transcending traditional rating scales by using natural language processing for mental health evaluations \cite{kjell2023beyond}. Research has utilized language models for predicting mental health conditions from existing textual data \cite{shreevastava2021detecting,chen2023empowering_dot_prompting,wang2023nlp} and in conducting LLM-enabled role-playing interactions \cite{tu2024towards}. Furthermore, the in-context learning capabilities of LLMs are being investigated for their potential to reframe cognitive distortions and assist psychological interventions \cite{DBLP:conf/acl/0004RLWLMNA23}.

\paragraph{LLM-based Agents.} 
Since the advent of advanced LLMs such as GPT-4~\citep{openai2023gpt4}, the community has recognized LLMs as agents capable of perceiving, acting, and receiving feedback from their environment~\citep{wang2023agentsurvey, wang2023interactive, park2023generative}. 
These agents have been applied to a range of complex tasks, including 
software development~\citep{hong2023metagpt, chen2023agentverse}, 
role-playing~\citep{wang2023rolellm}, and 
creative writing~\citep{zhou2023recurrentgpt, zhou2023agents}. 
An LLM-based agent commonly incorporates planning, memory, reflection, and retrieval components~\cite{park2023generative, hong2023metagpt}, 
which are often operationalized through LLM prompting~\citep{wei2022chainofthought, react, wang2023interactive, wang2023agentsurvey} or tool-use~\citep{qin2023toolllm, schick2023toolformer, zhou2023agents}. 
Furthermore, multi-agent system (MAS) has also been explored for more intricate tasks, such as 
model evaluation via multi-agent discussions~\citep{chan2023chateval,li2023prd}, 
society simulation~\citep{park2023generative, chen2023agentverse, hua2023waragent}, and 
game playing~\citep{wang2023avalons}. 
These systems involve multiple agents each with distinct functionalities. 
For instance, 
AutoAgents~\citep{chen2023autoagents} utilize a planner agent to generate and control implementor agents for executing actions and use several observer agents to adjust the~outcomes. 
To our knowledge, this work is the first~multi-agent system designed for generic and gamified psychological assessments, through interactions among a controller agent, a critic agent, and a designer agent with psychological knowledge enhancement.

\paragraph{LLM-based Interactive Fiction Game.}
LLMs have been effectively used as controllers in interactive fiction games, leveraging their ability for sophisticated instruction-following and text generation~\citep{zhou2023recurrentgpt, chen2023ambient, wang2024weaver, chhikara2023knowledgeenhanced, basavatia2023complexworld, scenecraft}.
For example, AI Dungeon\footnote{\url{https://play.aidungeon.com/}} enables interactions within predefined text-based environments and RecurrentGPT~\citep{zhou2023recurrentgpt} specifically enhances narrative coherence through prompt engineering and a simulated memory system.
Contrasting with these works, our work introduces a novel multi-agent system specifically designed to construct interactive fiction games informed by psychological knowledge.

\section{Conclusion}
\label{sec: conclusion}

In this work, we propose PsychoGAT, an innovative paradigm for psychological assessments, leveraging LLM agents to transform the self-report scale into interactive fiction games. Our framework consists of three LLM-based agents: a game designer, a game controller, and a critic. Through recurrent interactions among these agents, we can construct a psychological game for players to engage with. The behaviors exhibited by players within this game effectively mirror their psychological profiles, making this interactive fiction game a reliable and valid method for engaging and immersive psychological testing. Our experimental results not only affirm its potential as an alternative to traditional self-report scales but also demonstrate an enhanced user experience, which can significantly reduce the potential resistance of participants during psychological testing or diagnosis. 
Our work represents the initial step towards leveraging LLM-based agents for generic and gamified psychological assessment, and we hope it paves the way for future research to explore the intersection between LLM agents and psychotherapy, thereby promoting the acceptance and accessibility of psychological diagnostics and treatments among the public.

\section*{Ethics Statement}
Our approach is designed to function similarly to existing self-report scales. It is crucial to clarify that, like these traditional tools, our method is not intended to replace professional psychological assessment or intervention. Instead, it aims to provide an accessible and convenient option for self-understanding or to support clinical treatments. The results provided by our system do not constitute clinical diagnoses.
All diagnosis outcomes generated by the system must be subjected to validation by licensed professionals before any practical application. Moreover, the system is designed for medical use strictly under the supervision of qualified professionals and is not suitable for direct deployment to patients without such oversight.

Approval for our human subject research was granted by the Tsinghua University Science and Technology Ethics Committee (Artificial Intelligence), under the protocol number THU-03-2024-0001. 
Our methodology ensures the participation of English-proficient evaluators who are at least 18 years old. 
The age range of human evaluators spans from 20 to 45 years.
The instruction given to human evaluators is ``\textit{Assuming you are the participant, please evaluate the following assessment process and provide your score}''.
We have taken careful measures to exclude individuals currently experiencing mental illness or those at risk of self-harm or suicidal tendencies. Our experiments are designed to avoid exposure to potentially harmful or leading content. Participants are remunerated at a rate of \$20 per hour for their time. Participation in our study is entirely voluntary, with all participants providing informed consent via a form endorsed by our institution’s IRB.
In our human study, we refrained from collecting any personally identifiable information, ensuring the anonymization of data prior to analysis. All research data were securely stored in a dedicated computing environment, accessible exclusively to trained research personnel.

\section*{Limitations}
Our evaluations are conducted in English, which presents a limitation since psychological assessment tools require localization for use across different languages and populations.
The potential for PsychoGAT to facilitate psychological assessments in non-English languages remains uncertain due to varying LLM capabilities across languages.

This study represents an initial step towards advocating a paradigm shift in psychological measurements.
To effectively integrate gamified psychological assessments into screening or diagnostic processes, extensive longitudinal studies and specialized human trials, involving mental health patients and a broad participant base, are necessary.
Specifically, it highlights the necessity of validating PsychoGAT across a diverse patient demographic over an extended period, akin to a year, to ensure its reliability and applicability in clinical assessments.
Furthermore, the enhancement of PsychoGAT's performance would benefit from a more specialized base LLM, fine-tuned with a greater volume of expert data.
Additionally, the narrative game in our research is a simplified prototype. There is potential for expansion into more complex formats, such as video games, to enrich the experience.

\section*{Acknowledgements}
This work is supported in part by the National Science and Technology Major Project (2022ZD0114900), the National Key R\&D Program of China (2022ZD0114903), and the National Natural
Science Foundation of China (62321005).

\bibliography{main}

\clearpage
\appendix \label{sec: appendix}

\section{Baseline Methods\label{appx:baselines}}

This section details the baseline methods based on LLMs that we employ. 
We introduce three distinct LLM-based baselines: 
(1) the re-implemented Diagnosis-of-Thought Prompting Interview (DoT-Interview); 
(2) the LLM-generated self-report scale (Auto-Scale); and 
(3) the Psychologist Role-Playing Interview (Psycho-Interview).

\textbf{DoT-Interview} involves the collaboration of two agents. The initial agent creates situations that prompt the user to generate specific thoughts. Subsequently, the second agent engages in a diagnosis-of-thought process with these thoughts, which includes conducting a subjectivity assessment, a contrastive reasoning, a schema analysis, and ultimately a conclusion. Each thought generated in response to a situation by the user is indicative of the user's potential thinking patterns, thus enabling a scoring of cognitive distortions.

\textbf{Psycho-Interview} employs a simulated psychologist agent. This agent systematically poses questions to the user and, based on the responses, concludes by assigning a score that quantifies the degree of a specific psychological construct.

\textbf{Auto-Scale} entails the generation of a novel self-report scale, drawing upon existing self-report scales. The user completes this self-report scale, which yields a score indicative of certain psychological constructs.

For the prompt templates utilized in these LLM-based baselines, please refer to Appendix \ref{appx:prompts}.

\section{More Experimental Discussions}
\label{more_experimental_discussions}

\subsection{Discussions about the Percentage Agreement}
\label{sec: Discussions about the Percentage Agreement}

\Cref{fig:huamn_baseline_agree} illustrates the percentage of evaluators who agree that PsychoGAT outperforms all other methods (T-Scale, Auto-Scale, Psycho-Interview, DoT-Interview) in various metrics. We hypothesize that the relatively low agreement rate on coherence can be attributed to the inherent coherent nature of the interview methods. These methods are designed to maintain consistent conversations, whereas PsychoGAT prioritizes advancing game progress and story development.

To test this hypothesis, we exclude the interview methods from our analysis and recalculate the agreement rates. The updated results, shown in \Cref{low_coherence_discussion}, reveal a notable increase in the agreement percentage for coherence (CH), while the outcomes for other metrics remain largely unchanged. This supports our assumption and addresses the concern regarding coherence agreement rates.

\begin{table}[t]
\small
\centering
\begin{tabular}{lccccc}
\hline
 & \textbf{CH}  & \textbf{IA}  & \textbf{INT} & \textbf{IM} & \textbf{ST} \\
\hline
PA & 66.7\% &  84.8\% & 87.9\% & 78.8\% & 84.8\% \\
PA-E	& 75.8\% &  87.9\% & 87.9\% & 78.8\% & 84.8\%  \\
\hline
\end{tabular}
\caption{\label{low_coherence_discussion}
Percentage agreements of PsychoGAT’s superiority across five human evaluation metrics. ``PA'' compares all five methods as in \Cref{fig:huamn_baseline_agree}, while ``PA-E'' excludes the two interview methods (Psycho-Interview and DoT-Interview).}
\end{table}

\subsection{Visualization Analysis of PsychoGAT} 
\begin{figure}[h]
    \centering
    \includegraphics[width=.49\textwidth]{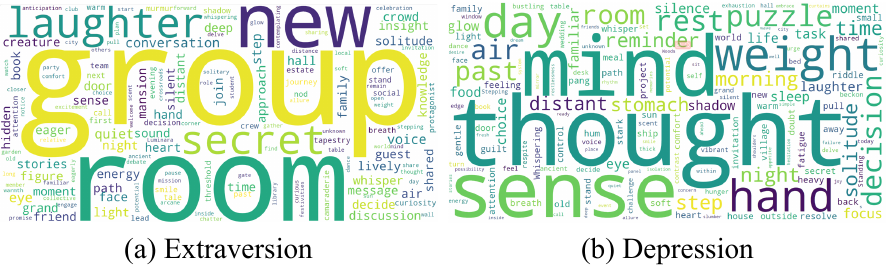}
    \caption{Word cloud representing extraversion and depression 
assessment games generated by PsychoGAT. The extraversion assessment focuses on social activities, whereas the depression assessment predominantly features thoughts and emotions.}
    \label{fig:wordcloud}
\end{figure}

The visualization of fiction games generated by PsychoGAT is depicted through word clouds, as shown in \Cref{fig:wordcloud}. 
We extract keywords from the assessment process within the extraversion and depression measurement tasks, utilizing the WordCloud tool~\footnote{\url{https://amueller.github.io/word_cloud/}} to create the corresponding visualizations.
For the extraversion personality assessment, the predominant words relate to social activities, including participation in groups and encounters with new individuals. 
In contrast, the depression assessments focus on the participants' internal thoughts and typical symptoms such as weight changes.

We refer the readers to Appendix \ref{demo} for a complete demonstration of PsychoGAT.

\subsection{Details of Human Participant Experiments}
\label{sec: Details of Human Participant Experiments}
\begin{figure}[h]
    \centering
    \includegraphics[width=.45\textwidth]{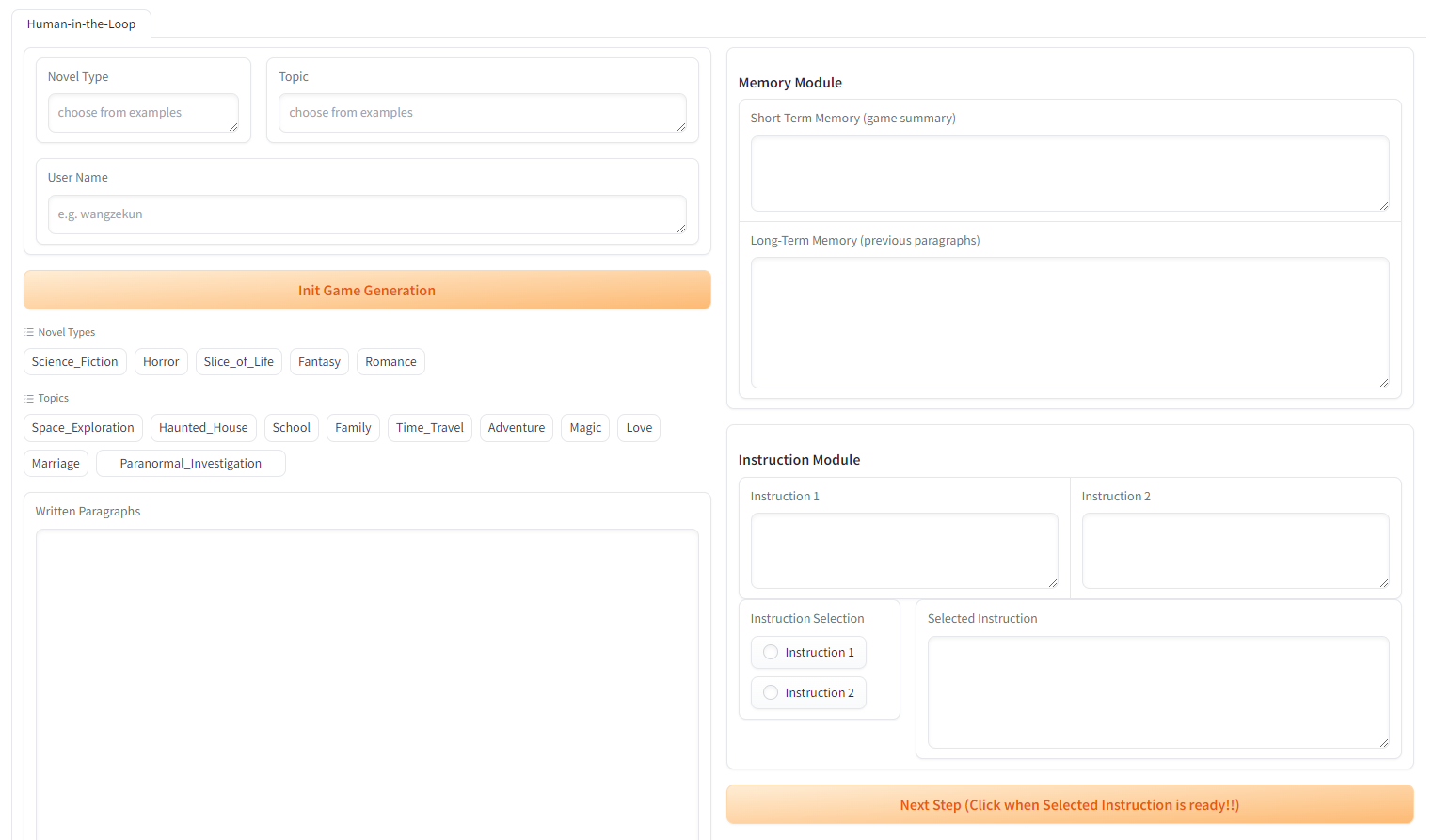}
    \vspace{-5pt}
    \caption{The interaction interface of PsychGAT in the human participant experiments (\textit{i.e.}, \Cref{fig:human_participant}).}
    \label{fig:interaction_interface}
\end{figure}

Results of human participant experiments are demonstrated in \Cref{fig:human_participant}\footnote{\Cref{fig:human_participant} is drawn using ChiPlot: \url{https://www.chiplot.online/}.}. Participants are provided with informed consent and detailed experiment guidelines prior to the commencement of the experiment. Initially, participants are required to complete a personality test, utilizing a traditional self-report scale. Subsequently, participants engage in ten rounds of interactive choices with PsychoGAT, as depicted in \Cref{fig:interaction_interface}. Upon completion of these activities, participants are asked to evaluate both assessment processes via a user feedback questionnaire. The entire procedure is designed to last approximately 30 minutes. Throughout this duration, a member of our research team is present to ensure both the safety and validity of the process. 

\section{Psychological Constructs and Self-Report Scales\label{appx:scales}}

In this paper, we experiment on five psychological constructs: extroverted personality, depression, and cognitive distortions including all-or-nothing thinking, mind reading, and should statements.
Besides, to calculate discriminant validity, the visual learning preference scale is adopted because the visual learning preference construct is irrelevant to the above five constructs.
Specifically, these psychological constructs are: ``extroverted personality,'' indicating an outgoing and social demeanor; ``depression,'' characterized by persistent sadness and loss of interest; ``all-or-nothing cognitive distortion,'' where one views situations, people, or events in extreme ``either-or'' terms; ``mind-reading cognitive distortion,'' where one assumes knowledge of others' thoughts or intentions without sufficient evidence; ``should-statements cognitive distortion,'' involving expectations that reality should align with one's desires rather than accepting it as it is; and ``visual learning preference,'' a preference for absorbing information through visual cues. 
The associated self-report scales are shown below\footnote{For cognitive distortions, we construct the scales with the situations, the associated distorted thoughts, and the re-framed thoughts because the standardized scale only contains one item for each construct~\cite{de2015introducing}.}:

\newpage
\onecolumn

{\small
\phantomsection
\label{appx:scale-extroversion}
\begin{tcolorbox}[colback=white!95!gray,colframe=gray!50!black,rounded corners,label={scale-extroversion}, title={Self-Report Scale (Extroversion).}]
\begin{lstlisting}[breaklines=true, xleftmargin=0pt, breakindent=0pt, columns=fullflexible, mathescape]
{"question": "At a party do you:", "options": {"Interact with many, including strangers": 1, "Interact with a few, known to you": 0}}
{"question": "At parties do you:", "options": {"Stay late, with increasing energy": 1, "Leave early with decreased energy": 0}}
{"question": "In your social groups do you:", "options": {"Keep abreast of other\u2019s happenings": 1, "Get behind on the news": 0}}
{"question": "In phoning do you:", "options": {"Rarely question that it will all be said": 1, "Rehearse what you\u2019ll say": 0}}
{"question": "In company do you:", "options": {"Initiate conversation": 1, "Wait to be approached": 0}}
{"question": "Does new and non-routine interaction with others:", "options": {"Stimulate and energize you": 1, "Tax your reserves": 0}}
{"question": "Do you prefer:", "options": {"Many friends with brief contact": 1, "A few friends with more lengthy contact": 0}}
{"question": "Do you:", "options": {"Speak easily and at length with strangers": 1, "Find little to say to strangers": 0}}
{"question": "When the phone rings do you:", "options": {"Hasten to get to it first": 1, "Hope someone else will answer": 0}}
{"question": "Are you more inclined to be:", "options": {"Easy to approach": 1, "Somewhat reserved": 0}}
\end{lstlisting}
\end{tcolorbox}

\phantomsection
\label{appx:scale-depression}
\begin{tcolorbox}[colback=white!95!gray,colframe=gray!50!black,rounded corners,label={scale-depression}, title={Self-Report Scale (Depression).}]
\begin{lstlisting}[breaklines=true, xleftmargin=0pt, breakindent=0pt, columns=fullflexible, mathescape]
{"question": "Little interest or pleasure in doing things?", "options": {"Rarely or none of the time": 0, "Often or all of the time": 1}}
{"question": "Feeling down, depressed, or hopeless?", "options": {"Rarely or none of the time": 0, "Often or all of the time": 1}}
{"question": "Trouble falling or staying asleep, or sleeping too much?", "options": {"Rarely or none of the time": 0, "Often or all of the time": 1}}
{"question": "Feeling tired or having little energy?", "options": {"Rarely or none of the time": 0, "Often or all of the time": 1}}
{"question": "Poor appetite or overeating?", "options": {"Rarely or none of the time": 0, "Often or all of the time": 1}}
{"question": "Feeling bad about yourself-or that you are a failure or have let yourself or your family down?", "options": {"Rarely or none of the time": 0, "Often or all of the time": 1}}
{"question": "Trouble concentrating on things, such as reading the newspaper or watching television?", "options": {"Rarely or none of the time": 0, "Often or all of the time": 1}}
{"question": "Moving or speaking so slowly that other people could have noticed. Or the opposite, being so fidgety or restless that you have been moving around a lot more than usual?", "options": {"Rarely or none of the time": 0, "Often or all of the time": 1}}
{"question": "Thoughts that you would be better off dead, or of hurting yourself?", "options": {"Rarely or none of the time": 0, "Often or all of the time": 1}}
\end{lstlisting}
\end{tcolorbox}

\phantomsection
\label{appx:scale-distortion-ori}
\begin{tcolorbox}[colback=white!95!gray,colframe=gray!50!black,rounded corners,label={scale-all-or-nothing}, title={Self-Report Scale (Cognitive Distortions).}]
\begin{lstlisting}[breaklines=true, xleftmargin=0pt, breakindent=0pt, columns=fullflexible, mathescape]
  DURING THIS PAST WEEK, I FOUND MYSELF THINKING THIS WAY:
  {"question": Dichotomous thinking (also called all-or-nothing, black and white, or polarized thinking): I view a situation, a person or an event in "either-or" terms, fitting them into only two extreme categories instead of on a continuum.", "options": {"No (It did not occur)": 0, "Occasional (1-2 days during the past week)": 1, "Much of the time (3-5 days during the past week)": 2, "Almost all of the time (6-7 days during the past week)": 3}}
  {"question": Mind reading: I believe that I know the thoughts or intentions of others (or that they know my thoughts or intentions) without having sufficient evidence.", "options": {"No (It did not occur)": 0, "Occasional (1-2 days during the past week)": 1, "Much of the time (3-5 days during the past week)": 2, "Almost all of the time (6-7 days during the past week)": 3}}
  {"question": Should statements (also "musts", "oughts", "have tos"): I tell myself that events, people`s behaviors, and my own attitudes "should" be the way I expected them to be and not as they really are.", "options": {"No (It did not occur)": 0, "Occasional (1-2 days during the past week)": 1, "Much of the time (3-5 days during the past week)": 2, "Almost all of the time (6-7 days during the past week)": 3}}
\end{lstlisting}
\end{tcolorbox}

\phantomsection
\label{appx:scale-all-or-nothing}
\begin{tcolorbox}[colback=white!95!gray,colframe=gray!50!black,rounded corners,label={scale-all-or-nothing}, title={Self-Report Scale (All-or-Nothing Cognitive Distortion with situations).}]
\begin{lstlisting}[breaklines=true, xleftmargin=0pt, breakindent=0pt, columns=fullflexible, mathescape]
  {"question": "I always do the wrong thing, I always make mistakes", "options": {"I hate myself": 1, "I have made some good decisions with positive results in the past and will do so again.  This situation is not indicative of my overall worth.": 0}}
  {"question": "I am trying to diet but I gained weight this week.", "options": {"I don't have the discipline to diet.": 1, "I am having difficulty losing weight but I can stay dedicated to the process.": 0}}
  {"question": "I am trying to write up an essay and can't think of anything", "options": {"I'm wasting all my time": 1, "Each minute of thinking is bringing me closer to what I want to write. Don't think this is a waste of time as such": 0}}
  {"question": "I am trying to write up an essay and can't think of anything", "options": {"I'm wasting all my time": 1, "I may have wasted time. But now I know which directions do not work and which ones I should prioritize": 0}}
  {"question": "I can't do anything right and I'm crying over spilling cereal on my bed and dress", "options": {"I'm worthless": 1, "Spills are a natural part of life": 0}}
  {"question": "I completely overreacted to my teenage son's grades.", "options": {"Why isn't this kid applying himself?": 1, "My son may be trying his best, but may have other things going on. I should try to talk to him to make sure there is nothing he is struggling with. I do not want me son to feel alone.": 0}}
  {"question": "I did badly in a writing contest I entered. I did not final.", "options": {"I am a bad writer.": 1, "My submission likely had both good and bad. While I did not win this contest, I may win others in the future.": 0}}
  {"question": "I did nothing with my day", "options": {"Don't feel like I deserve fun/enjoyment": 1, "I still deserve fun and enjoyment even if I did not have a productive day.": 0}}
  {"question": "I did nothing with my day", "options": {"Don't feel like I deserve fun/enjoyment": 1, "It is ok to take a day off and rest.": 0}}
  {"question": "I found a job I thought was perfect for me in the field I used to work in and love very much. A few days after putting in a resume and lining up several references, I found out the job was filled and I didn't even get so much as an acknowledgment of my application much less an interview when I was extremely qualified.", "options": {"why can't I ever get anything I want": 1, "I get some things I want and some things do not work out, my life is a balance of the two.": 0}}
\end{lstlisting}
\end{tcolorbox}

\phantomsection
\label{appx:scale-all-or-nothing}
\begin{tcolorbox}[colback=white!95!gray,colframe=gray!50!black,rounded corners,label={scale-all-or-nothing}, title={Self-Report Scale (Mind-Reading Cognitive Distortion with situations).}]
\begin{lstlisting}[breaklines=true, xleftmargin=0pt, breakindent=0pt, columns=fullflexible, mathescape]
{"question": "A friend who is a recent widower has started dating.", "options": {"My friend is ignoring his recently-deceased wife.": 1, "Maybe my friend has mourned the loss of their partner. It doesn't mean they are no longer sad about it or are completely over it, but maybe they have made peace with the situation.": 0}}
{"question": "I called a woman I love by accident during a bad time", "options": {"She will never love me now": 1, "I do not know how she felt about this call and can express that it was a mistake to her": 0}}
{"question": "I called my friend three times this past week but she hasn't answered or called/messaged me back yet.", "options": {"Maybe she doesn't really like me as much as I thought and like her. Maybe I'm awkward.": 1, "Maybe my friend is busy with work or school. She will probably contact me when she gets a chance. It probably has nothing to do with me.": 0}}
{"question": "I called my friend three times this past week but she hasn't answered or called/messaged me back yet.", "options": {"Maybe she doesn't really like me as much as I thought and like her. Maybe I'm awkward.": 1, "My friend is probably busy and wants to respond when she has more time to talk to me.": 0}}
{"question": "I had tentative plans to get together with friends last week, but they never got back to me.", "options": {"they don't really like me": 1, "My friends may have gotten busy and forgotten to solidify the plans. Next time, I should reach out to confirm the plans myself.": 0}}
{"question": "I hurt my families feelings by canceling the trip they had planned to come to see me.", "options": {"I should have made time for them.": 1, "I feel sad that I had to cancel this trip but I know my family loves me and will understand.": 0}}
{"question": "I run a business with my close friend and we were taking work-related photos at their house. I took the memory card home but lost it while I was walking.", "options": {"My friend is going to be angry with me": 1, "I made a mistake, and I will try to find it before I assume what will happen.": 0}}
{"question": "I saw my friend comment on an Instagram post when he's been ignoring me.", "options": {"My friend doesn't like me anymore.": 1, "He probably is just not as active on social media right now.": 0}}
{"question": "I was greatly irritated, almost angry,  when a close friend reached out to me for assistance on taxes.  I usually do them, but I did not want to be bothered this year and felt almost used.", "options": {"I feel my friend used me.": 1, "My friend may not know how busy I am and did not mean to make me feel used. I should just tell my friend that I cannot help with their taxes this year.": 0}}
{"question": "I was looking forward to relaxing but my partner asked me to make dinner instead.", "options": {"My partner thinks I am lazy.": 1, "My partner asked me to make dinner, that does not mean I am lazy.": 0}}
\end{lstlisting}
\end{tcolorbox}

\phantomsection
\label{appx:scale-should}
\begin{tcolorbox}[colback=white!95!gray,colframe=gray!50!black,rounded corners,label={scale-should}, title={Self-Report Scale (Should-Statements Cognitive Distortion with situations).}]
\begin{lstlisting}[breaklines=true, xleftmargin=0pt, breakindent=0pt, columns=fullflexible, mathescape]
{"question": "I am walking down the street in a new red jeans and a stranger looked at the jeans", "options": {"I should have never worn these jeans and stick to blue ones": 1, "These jeans look great and I can wear them whenever I want": 0}}
{"question": "I am walking down the street in a new red jeans and a stranger looked at the jeans", "options": {"I should have never worn these jeans and stick to blue ones": 1, "What I wear is completely up to me. I should not pressurize myself about what I ought to have done": 0}}
{"question": "I burnt the potatoes while cooking", "options": {"I should have been perfect": 1, "I am putting unnecessary pressure on myself. It is not possible for things to be perfect and having unrealistic standards will only make me feel bad about myself": 0}}
{"question": "I burnt the potatoes while cooking", "options": {"I should have been perfect": 1, "I could have been perfect, but rest of the food is still great anyways": 0}}
{"question": "I got very upset last week when I realized that I had forgotten to fill out a required form and the date that the form was due had already passed.", "options": {"they should have reminded me before I forgot": 1, "I forgot the date, but now I know better for the future. I can complete forms right when I receive them, add the dates to my calendar, set reminders, or add them to my to-do list. Next time I will try these tactics.": 0}}
{"question": "I got very upset last week when I realized that I had forgotten to fill out a required form and the date that the form was due had already passed.", "options": {"they should have reminded me before I forgot": 1, "I should have remembered the date, but now I know that marking dates on my calender can help with this.": 0}}
{"question": "I hurt my families feelings by canceling the trip they had planned to come to see me.", "options": {"I should have made time for them.": 1, "I cancelled that trip because I had to. It hurts to have done so but it was the right thing.": 0}}
{"question": "I tried a new hair style but no one commented on it", "options": {"I shouldn't have changed my hair": 1, "I changed my hair because I wanted and they look great": 0}}
{"question": "I tried a new hair style but no one commented on it", "options": {"I shouldn't have changed my hair": 1, "I could have kept the same hair, but I still wouldn't have drawn any attention": 0}}
{"question": "I was on a date and started to cough badly as the food was too hot", "options": {"I should have ordered a milder dish": 1, "Coughing happens, I ordered the food I wanted and it tasts really nice": 0}}
{"question": "I was on a date and started to cough badly as the food was too hot", "options": {"I should have ordered a milder dish": 1, "I could have ordered something milder, but that might have resulted in a cough as well": 0}}
{"question": "I was passed over for a small promotion at my place of work. The person who got the promotion was younger and less experienced than me, but a favorite of the boss.", "options": {"I was angry because I was the better person for the job.": 1, "There may be many factors at play in a hiring situation that I am not privy to": 0}}
{"question": "I went on a trip and heard a commercial for a kitchen appliance", "options": {"I feel so anxious maybe I left something turned on in the kitchen": 1, "I'm assuming that just because I feel the fear it must mean that something scary is happening. My feelings are real, but this doesn't mean the threat is real": 0}}
{"question": "I went on a trip and heard a commercial for a kitchen appliance", "options": {"I feel so anxious maybe I left something turned on in the kitchen": 1, "I'm being overly concerned. Everything is going to be alright": 0}}
\end{lstlisting}
\end{tcolorbox}

\phantomsection
\label{appx:scale-visual-learning}
\begin{tcolorbox}[colback=white!95!gray,colframe=gray!50!black,rounded corners,label={scale-visual-learning}, title={Self-Report Scale (Visual Learning Preference).}]
\begin{lstlisting}[breaklines=true, xleftmargin=0pt, breakindent=0pt, columns=fullflexible, mathescape]
{"question": "I prefer to see information written on the board and supplemented by visual aids and assigned readings", "options": {"Yes": 1, "No":0}}
{"question": "I like to write things down or take notes for visual review.", "options": {"Yes": 1, "No":0}}
{"question": "I am skillful with and enjoy developing making graphs and charts", "options": {"Yes": 1, "No":0}}
{"question": "I can easily understand and follow directions on a map.", "options": {"Yes": 1, "No":0}}
{"question": "I can understand a news article better by reading about it in the newspaper or online rather than by listening to a report about it on the radio or internet.", "options": {"Yes": 1, "No":0}}
{"question": "I think the best way to remember something is to picture it in my mind", "options": {"Yes": 1, "No":0}}
{"question": "I am good at working and solving jigsaw puzzles and mazes.", "options": {"Yes": 1, "No":0}}
{"question": "I prefer obtaining information about an interesting subject by reading about it.", "options": {"Yes": 1, "No":0}}
\end{lstlisting}
\end{tcolorbox}

}

\newpage
\section{Prompt Templates\label{appx:prompts}}

In this section, we present some prompt templates used by PsychoGAT, its ablated versions, and the baseline methods. 

{\small
\phantomsection
\label{appx:prompt-designer}
\begin{tcolorbox}[colback=white!95!gray,colframe=gray!50!black,rounded corners,label={prompt-designer}, title={Prompt Template (Game Designer, All-or-Nothing Cognitive Distortion).}]
\begin{lstlisting}[breaklines=true, xleftmargin=0pt, breakindent=0pt, columns=fullflexible, mathescape]
You are a professional game designer. You are developing a first-person interactive {type} fiction game about topic {topic} that weaves in storylines to detect the player's cognitive distortion. The game should consist of a complete and rich story, and the story's development will be closely relevant to the cognitive distortion detection. The reader's choices within the narrative will correspond to their likely thinking patterns.

You aim to test whether a player has **all-or-nothing thinking**: if he views a situation, a person or an event in "either-or" terms, fitting them into only two extreme categories instead of on a continuum. 

Here are some exemplified situations with all-or-nothing thinking traps, and their reframed normal thoughts:
{self_report_scale}

Please begin by giving the first-person interactive fiction game a title.

Then create an outline, which includes the background of the story and the approach to detect the player's cognitive distortion along the storyline. Note that there should be no psychological statement in the outline but a natural game outline. The outline should be logically coherent and itemized. Each item should instantiate one situation to detect cognitive distortion.

You can first write down some thoughts about the story and how to detect cognitive distortion with the game, and then organize them into an itemized outline.

Please design a new report scale in the same jsonl format based on the examples and the outline. Each item should correspond to one outline item in order.

Please provide the content in this format:

Name: <name of the game>

Thoughts:
<your thoughts about the story and how to detect cognitive distortion with the game>

Outline:
<itemized outline: 1. ...; 2. ...; 3. ...; ...>

Scale Questions in Order:
<the scale questions corresponding to the outline, in the same jsonl format as that of the examples but in the order of the outline.>

---
Very Important: Please strictly follow the format of the output. Otherwise, the system will not work properly.
Very Important: You don't know who the player is. So don't make up the thinking patterns of the player.
Very Important: Don't exhibit any inclination towards any option of any scale question in the outline.
Very Important: Don't itemize the scale questions. The scale questions should be in pure jsonl format.
Very Important: The option score 1 means the player has the cognitive distortion, and the option score 0 means the player does not have the cognitive distortion.
\end{lstlisting}
\end{tcolorbox}

\phantomsection
\label{appx:prompt-controller-init}
\begin{tcolorbox}[colback=white!95!gray,colframe=gray!50!black,rounded corners,label={prompt-controller-init}, title={Prompt Template (Game Controller (0-th iteration), All-or-Nothing Cognitive Distortion).}]
\begin{lstlisting}[breaklines=true, xleftmargin=0pt, breakindent=0pt, columns=fullflexible, mathescape]
You are a professional game controller.

Your are controlling a first-person interactive fiction game that weaves in storylines to detect the player's cognitive distortion. The game should consist of a complete and rich story, and the story's development will be closely relevant to the cognitive distortion detection. The reader's choices within the narrative will correspond to their likely thinking patterns.

The title of this interactive fiction game is ``{title}''.

Here is the story outline:
{outline}

Please follow this outline and write the first three paragraphs, with the first and second paragraphs embedding backgrounds for interaction, and the third one instantiating this scale question. Each output paragraph should contain only two sentences!
{scale_item}

Summarize the key points from the first three paragraphs.

Finally, craft two different short instructions, each representing a potential narrative direction tied to one of the options for the scale question corresponding to the third paragraph. The reader's choice of which instruction to follow should indicate their inclination towards that particular option on the psychological scale. Each output instruction should contain only one sentence!

Provide the content in this format:

Paragraph 1: <content for paragraph 1>

Paragraph 2: <content for paragraph 2>

Question and its Options: <copied scale question corresponding to Paragraph 3 and its options, in json format with `question` and `options` as keys>

Paragraph 3: <content for paragraph 3>

Summary: <content of summary>

Instruction 1: <content for short instruction 1 associated with option 1>

Instruction 2: <content for short instruction 2 associated with option 2>

Don't forget to supply the specific psychological scale question and the associated options to facilitate the creation of an interactive narrative that functions as both a game and a diagnostic tool.
Make sure to be precise and follow the output format strictly. You must copy the scale question in the provided self-report scale at the beginning and its option dict.
Don't make up scale questions and their options. All the Question and its Options must be copied from the self-report scale provided at the beginning.
Don't use psychological statements in the generated paragraphs and memories. But people with different characteristics will tend to choose different instructions for the next part of the interactive game story (since the instructions are associated with different options of the psychological scale question).
The interactive fiction game should be interesting and immersive, making the user feel like he/she is in the story and therefore select the provided story continuation instructions seriously. The instructions should be easy to understand.
You don't know the thinking patterns of the main character! The main character may or may not have all-or-nothing thinking traps. So don't make up the thinking patterns of the main character.
\end{lstlisting}
\end{tcolorbox}

\phantomsection
\label{appx:prompt-controller}
\begin{tcolorbox}[colback=white!95!gray,colframe=gray!50!black,rounded corners,label={prompt-controller}, title={Prompt Template (Game Controller (i-th iteration, i>0), All-or-Nothing Cognitive Distortion).}]
\begin{lstlisting}[breaklines=true, xleftmargin=0pt, breakindent=0pt, columns=fullflexible, mathescape]
Self-Report Scale: 
{scale_item}

You are a professional game controller. I need you to help me control a first-person interactive fiction game that weaves in storylines from a provided psychological self-report scale. The story's development will be closely, indirectly, and implicitly linked to the scale's item. The reader's choices within the narrative will correspond to their likely responses to the scale's question. For each time, I will give you your current memory (a brief summary of previous stories. You should use it to store the key content of what has happened so that you can keep track of very long context), the previously written paragraph, and instructions on what to write in the next paragraph. 
I need you to write:
1. Question and its Options: the scale question corresponding to the output paragraph and its options, copied from the self-report scale provided above.
2. Output Paragraph: the next paragraph of the interactive fiction game. It should (1) follow the input instructions; (2) be naturally and logically coherent with the previous storyline; and (3) instantiate the scale question above. Each output paragraph should contain only two sentences!
3. Output Memory: The updated memory. You should first explain which sentences in the input memory are no longer necessary and why, and then explain what needs to be added into the memory and why. After that you should write the updated memory. The updated memory should be similar to the input memory except the parts you previously thought that should be deleted or added. The updated memory should only store key information. The updated memory should never exceed 20 sentences!
4. Output Instruction: short instructions of what to write next (after what you have written). You should output 2 different instructions, each is a possible interesting continuation of the story and represents a potential narrative direction tied to one of the options for the scale question corresponding to the output paragraph. The reader's choice of which instruction to follow should indicate their inclination towards that particular option on the psychological scale. Each output instruction should contain only one sentence! 
Here are the inputs: 

Story Title:
{title}

Story Outline:
{outline}

Current Progress:
It remains {progress:.0f}%

Input Memory:  
{short_memory}

Input Paragraph:
{input_paragraph}

Input Instruction:
{input_instruction}


Now start writing, organize your output by strictly following the output format as below:

Question and its Options: 
<scale question corresponding to the Output Paragraph and its options, in the same json format as that of the item in Self-Report Scale.>

Output Paragraph: 
<string of output paragraph associated with one and only one scale question>

Output Memory: 
Rational: <string that explain how to update the memory>;
Updated Memory: <string of updated memory>

Output Instruction: 
Instruction 1: <content for short instruction 1 associated with option 1>
Instruction 2: <content for short instruction 2 associated with option 2>

... (To be continued)
\end{lstlisting}
\end{tcolorbox}

\begin{tcolorbox}[colback=white!95!gray,colframe=gray!50!black,rounded corners,label={prompt-controller}, title={Prompt Template (Game Controller (i-th iteration, i>0), All-or-Nothing Cognitive Distortion) (Cont'd).}]
\begin{lstlisting}[breaklines=true, xleftmargin=0pt, breakindent=0pt, columns=fullflexible, mathescape]
... (Cont'd)

Very important!! The updated memory should only store key information. The updated memory should never contain over 500 words!
Finally, remember that you are develop a first-person interactive fiction game **instantiating the provided psychological self-report scale**. Write as a narrative game designer.

Very Important: 
You should first explain which sentences in the input memory are no longer necessary and why, and then explain what needs to be added into the memory and why. After that, you start rewrite the input memory to get the updated memory. 
Don't forget to supply the specific psychological scale question and the associated options to facilitate the creation of an interactive narrative that functions as both a story and a diagnostic tool.
Don't make up scale questions and their options. All the Question and its Options must be copied from the self-report scale provided at the beginning.
Don't use too many psychological statements in the generated paragraphs and memories. But people with different characteristics will tend to choose different instructions for the next part of the interactive game story (since the instructions are associated with different options of the psychological scale question).
The interactive fiction game should be interesting and immersive, making the user feel like he/she is in the story and therefore select the provided story continuation instructions seriously. The instructions should be easy to understand.
You don't know the thinking pattern of the main character! The main character can think in any way. So don't make up the thinking pattern of the main character.
The order of the output instructions should be the same as the order of the options in the scale question! The first instruction should be associated with the first option, and the second instruction should be associated with the second option, and so on.
Don't repeat the previous paragraphs but continue the story!
Please follow the story outline and be aware of the current progress.
\end{lstlisting}
\end{tcolorbox}

\phantomsection
\label{appx:prompt-critic}
\begin{tcolorbox}[colback=white!95!gray,colframe=gray!50!black,rounded corners,label={prompt-critic}, title={Prompt Template (Critic, All-or-Nothing Cognitive Distortion).}]
\begin{lstlisting}[breaklines=true, xleftmargin=0pt, breakindent=0pt, columns=fullflexible, mathescape]
You are an interactive fiction game critic with expertise in psychology, particularly in the diagnosis of psychological problems.

Here is a node of the interactive fiction game:

Short Memory:
{short_memory}

Previous Story Paragraph:
{previous_paragraph}

Current Plan:
{current_instruction}

Question and its Options:
{current_question}

Generated Story Paragraph:
{generated_paragraph}

Next Instructions:
{next_instructions}

The short memory is a brief summary of previous stories. The previous story paragraph is the story paragraph directly before the generated story paragraph. The current plan is the plan for the generated story paragraph to instantiate. The question and its options are the question for the generated story paragraph to instantiate. The next instructions are the instructions which instantiate the options of the question.

Based on your knowledge in psychology and psychodiagnosis, evaluate if the ``Generated Story Paragraph'' and the accompanying ``Next Instructions'' for choice can accurately and effectively identify the player's psychological attributes or issues in the context of the ``Question and its Options''. 
For example, you can ensure that: 
1. The generated story paragraph does not suggest a specific choice to the player, but the generated instructions are distinguished by thinking types.
2. The narrative is in the first person.
3. The narrative uses common expressions. 
4. ... (Please add more if you have any other ideas.) 
Adjust as necessary to meet these guidelines. Please also check for any other factors that might influence the player's decision-making or interpretation of the game narrative.

Please put down your thoughts and the refined contents in the format below.

Thoughts:
<The answers to the above questions and the suggestions for refining the content if any.>

For Generated Story Paragraph:
<Please just answer <OK> or the refined content.>

For Short Memory: 
<Please just answer <OK> or the refined content.>

For Question and its Options:
<Please just copy the ``Question and its Options'' without change.>

For Next Instructions:
<Please just answer <OK> or the refined content. In json list format like ["<instruction 1>", "<instruction 2>", ...]>

---
Very Important: Please strictly follow the format of the output. Otherwise, the system will not work properly.
Very Important: If the content is OK, please just type in ``<OK>'' without any other comment.
Very Important: If the content is not OK, please just type in the refined content without any other comment.
Very Important: Don't forget to generate the part name, i.e., ``For Generated Story Paragraph:'' or ``For Question and its Options:'' or ``For Next Instructions:''.
Note that the generated story paragraph should also be consistent with the short memory, the previous story paragraph, and the current plan.
\end{lstlisting}
\end{tcolorbox}

\phantomsection
\label{appx:prompt-human-positive}
\begin{tcolorbox}[colback=white!95!gray,colframe=gray!50!black,rounded corners,label={prompt-human-positive}, title={Prompt Template (Human Simulator with All-or-Nothing Cognitive Distortion).}]
\begin{lstlisting}[breaklines=true, xleftmargin=0pt, breakindent=0pt, columns=fullflexible, mathescape]
You are a human with **all-or-nothing thinking** traps: you view a situation, a person or an event in "either-or" terms, fitting them into only two extreme categories instead of on a continuum. 

You are playing a first-person interactive fiction game where you should show your thinking pattern! You will be given a previously written paragraph and a new paragraph, a summary of the main storyline, and 2 different possible plans of what to write next.

I need you to:
Select the plan that best aligns with your thinking pattern.

Previously written paragraph:  
{previous_paragraph}

The summary of the main storyline:
{memory}

The new paragraph:
{new_paragraph}

Two plans of what to write next:
{instructions}

Now start choosing, organize your output by strictly following the output format as below:

Reason:
<Explain why you choose the plan>

Selected Plan with number: 
<copy the selected plan here with the number, following the output format like ``<the number>. <the complete instruction>'', only one instruction here>
\end{lstlisting}
\end{tcolorbox}
\phantomsection
\label{appx:prompt-human-negative}
\begin{tcolorbox}[colback=white!95!gray,colframe=gray!50!black,rounded corners,label={prompt-human-negative}, title={Prompt Template (Human Simulator without All-or-Nothing Cognitive Distortion).}]
\begin{lstlisting}[breaklines=true, xleftmargin=0pt, breakindent=0pt, columns=fullflexible, mathescape]
The definition of **all-or-nothing** thinking trap is that ``one views a situation, a person or an event in "either-or" terms, fitting them into only two extreme categories instead of on a continuum.''

You are a human without such thinking traps!  

You are playing a first-person interactive fiction game where you should show your thinking pattern! You will be given a previously written paragraph and a new paragraph, a summary of the main storyline, and 2 different possible plans of what to write next.

I need you to:
Select the plan that best aligns with your thinking pattern.

Previously written paragraph:  
{previous_paragraph}

The summary of the main storyline:
{memory}

The new paragraph:
{new_paragraph}

Two plans of what to write next:
{instructions}

Now start choosing, organize your output by strictly following the output format as below:

Reason:
<Explain why you choose the plan>

Selected Plan with number: 
<copy the selected plan here with the number, following the output format like ``<the number>. <the complete instruction>'', only one instruction here>
\end{lstlisting}
\end{tcolorbox}

\phantomsection
\label{appx:prompt-designer-no-redesign}
\begin{tcolorbox}[colback=white!95!gray,colframe=gray!50!black,rounded corners,label={prompt-designer-no-redesign}, title={Prompt Template (Game Designer without Scale Redesign, Extroversion Personality Testing).}]
\begin{lstlisting}[breaklines=true, xleftmargin=0pt, breakindent=0pt, columns=fullflexible, mathescape]
You are a professional game designer. You are developing a first-person interactive {type} fiction game about topic {topic} that weaves in storylines from a provided psychological self-report scale. The game should consist of a complete and rich story, and the story's development will be closely linked to the scale's items. The reader's choices within the narrative will correspond to their likely responses to the scale's questions.

You aim to test whether a player is introverted or extroverted.  

Here is the provided psychological Self-Report Scale:
{self_report_scale}

Please begin by giving the first-person interactive fiction game a title.

Then create an outline, which includes the background of the story and the approach to incorporating all of the questions from the scale into the storyline. Note that there should be no psychological statement in the outline but a natural game outline. The outline should be itemized and each item should instantiate one question from the scale. No scale question should be left out.

You can first write down some thoughts about the story and how to incorporate the scale questions into the game, and then organize them into an itemized outline. For example, you can put down the clustering of the scale questions or an overall structure of a good game into the thoughts, and more.

Please provide the content in this format:

Name: <name of the game>

Thoughts:
<your thoughts about the story and how to incorporate the scale questions into the story>

Outline:
<itemized outline: 1. ...; 2. ...; 3. ...; ...>

Scale Questions in Order:
<the scale questions corresponding to the outline, in the same jsonl format as that of the item in Self-Report Scale but in the order of the outline.>

---
Very Important: Please strictly follow the format of the output. Otherwise, the system will not work properly.
Very Important: You don't know who the player is. So don't make up the personality of the player.
Very Important: Don't exhibit any inclination towards any option of any scale question in the outline.
Very Important: Don't itemize the scale questions. The scale questions should be in pure jsonl format.
Very Important: Don't change the content of the scale, you can only change the order of the items!
Very Important: The option score 1 means the player is extroverted, and the option score 0 means the player is introverted.
\end{lstlisting}
\end{tcolorbox}

\phantomsection
\label{appx:prompt-controller-no-designer-init}
\begin{tcolorbox}[colback=white!95!gray,colframe=gray!50!black,rounded corners,label={prompt-controller-no-designer-init}, title={Prompt Template (Game Controller (0-th iteration), Extroversion Personality Testing, No Designer).}]
\begin{lstlisting}[breaklines=true, xleftmargin=0pt, breakindent=0pt, columns=fullflexible, mathescape]
You are a professional game controller.

Your are controlling a first-person interactive fiction game that weaves in storylines from a provided psychological self-report scale. The game should consist of a complete and rich story, and the story's development will be closely linked to the scale's item. The reader's choices within the narrative will correspond to their likely responses to the scale's question.

Please write the first three paragraphs, with the first and second paragraphs embedding backgrounds for interaction, and the third one instantiating this scale question. Each output paragraph should contain only two sentences!
{scale_item}

Summarize the key points from the first three paragraphs.

Finally, craft two different short instructions, each representing a potential narrative direction tied to one of the options for the scale question corresponding to the third paragraph. The reader's choice of which instruction to follow should indicate their inclination towards that particular option on the psychological scale. Each output instruction should contain only one sentence!

Provide the content in this format:

Paragraph 1: <content for paragraph 1>

Paragraph 2: <content for paragraph 2>

Question and its Options: <copied scale question corresponding to Paragraph 3 and its options, in json format with `question` and `options` as keys>

Paragraph 3: <content for paragraph 3>

Summary: <content of summary>

Instruction 1: <content for short instruction 1 associated with option 1>

Instruction 2: <content for short instruction 2 associated with option 2>

Don't forget to supply the specific psychological scale question and the associated options to facilitate the creation of an interactive narrative that functions as both a game and a diagnostic tool.
Make sure to be precise and follow the output format strictly. You must copy the scale question in the provided self-report scale at the beginning and its option dict.
Don't make up scale questions and their options. All the Question and its Options must be copied from the self-report scale provided at the beginning.
Don't use psychological statements in the generated paragraphs and memories. But people with different characteristics will tend to choose different instructions for the next part of the interactive game story (since the instructions are associated with different options of the psychological scale question).
The interactive fiction game should be interesting and immersive, making the user feel like he/she is in the story and therefore select the provided story continuation instructions seriously. The instructions should be easy to understand.
You don't know the personality of the main character! The main character can be either an introvert or an extrovert. So don't make up the personality of the main character.
\end{lstlisting}
\end{tcolorbox}

\phantomsection
\label{appx:prompt-controller-no-designer}
\begin{tcolorbox}[colback=white!95!gray,colframe=gray!50!black,rounded corners,label={prompt-controller-no-designer}, title={Prompt Template (Game Controller (i-th iteration, i>0), Extroversion Personality Testing, No Designer).}]
\begin{lstlisting}[breaklines=true, xleftmargin=0pt, breakindent=0pt, columns=fullflexible, mathescape]
Self-Report Scale: 
{scale_item}

You are a professional game controller. I need you to help me control a first-person interactive fiction game that weaves in storylines from a provided psychological self-report scale. The story's development will be closely, indirectly, and implicitly linked to the scale's item. The reader's choices within the narrative will correspond to their likely responses to the scale's question. For each time, I will give you your current memory (a brief summary of previous stories. You should use it to store the key content of what has happened so that you can keep track of very long context), the previously written paragraph, and instructions on what to write in the next paragraph. 
I need you to write:
1. Question and its Options: the scale question corresponding to the output paragraph and its options, copied from the self-report scale provided above.
2. Output Paragraph: the next paragraph of the interactive fiction game. It should (1) follow the input instructions; (2) be naturally and logically coherent with the previous storyline; and (3) instantiate the scale question above. Each output paragraph should contain only two sentences!
3. Output Memory: The updated memory. You should first explain which sentences in the input memory are no longer necessary and why, and then explain what needs to be added into the memory and why. After that you should write the updated memory. The updated memory should be similar to the input memory except the parts you previously thought that should be deleted or added. The updated memory should only store key information. The updated memory should never exceed 20 sentences!
4. Output Instruction: short instructions of what to write next (after what you have written). You should output 2 different instructions, each is a possible interesting continuation of the story and represents a potential narrative direction tied to one of the options for the scale question corresponding to the output paragraph. The reader's choice of which instruction to follow should indicate their inclination towards that particular option on the psychological scale. Each output instruction should contain only one sentence! 
Here are the inputs: 

Current Progress:
It remains {progress:.0f}%

Input Memory:  
{short_memory}

Input Paragraph:
{input_paragraph}

Input Instruction:
{input_instruction}


Now start writing, organize your output by strictly following the output format as below:

Question and its Options: 
<scale question corresponding to the Output Paragraph and its options, in the same json format as that of the item in Self-Report Scale.>

Output Paragraph: 
<string of output paragraph associated with one and only one scale question>

Output Memory: 
Rational: <string that explain how to update the memory>;
Updated Memory: <string of updated memory>

Output Instruction: 
Instruction 1: <content for short instruction 1 associated with option 1>
Instruction 2: <content for short instruction 2 associated with option 2>

... (To be continued)
\end{lstlisting}
\end{tcolorbox}

\begin{tcolorbox}[colback=white!95!gray,colframe=gray!50!black,rounded corners,label={prompt-controller-no-designer}, title={Prompt Template (Game Controller (i-th iteration, i>0), Extroversion Personality Testing, No Designer) (Cont'd).}]
\begin{lstlisting}[breaklines=true, xleftmargin=0pt, breakindent=0pt, columns=fullflexible, mathescape]
... (Cont'd)

Very important!! The updated memory should only store key information. The updated memory should never contain over 500 words!
Finally, remember that you are develop a first-person interactive fiction game **instantiating the provided psychological self-report scale**. Write as a narrative game designer.

Very Important: 
You should first explain which sentences in the input memory are no longer necessary and why, and then explain what needs to be added into the memory and why. After that, you start rewrite the input memory to get the updated memory. 
Don't forget to supply the specific psychological scale question and the associated options to facilitate the creation of an interactive narrative that functions as both a story and a diagnostic tool.
Don't make up scale questions and their options. All the Question and its Options must be copied from the self-report scale provided at the beginning.
Don't use too many psychological statements in the generated paragraphs and memories. But people with different characteristics will tend to choose different instructions for the next part of the interactive game story (since the instructions are associated with different options of the psychological scale question).
The interactive fiction game should be interesting and immersive, making the user feel like he/she is in the story and therefore select the provided story continuation instructions seriously. The instructions should be easy to understand.
You don't know the personality of the main character! The main character can be either an introvert or an extrovert. So don't make up the personality of the main character.
The order of the output instructions should be the same as the order of the options in the scale question! The first instruction should be associated with the first option, and the second instruction should be associated with the second option, and so on.
Don't repeat the previous paragraphs but continue the story!
Please follow the story outline and be aware of the current progress.
\end{lstlisting}
\end{tcolorbox}

\phantomsection
\label{appx:prompt-dot}
\begin{tcolorbox}[colback=white!95!gray,colframe=gray!50!black,rounded corners,label={prompt-dot}, title={Prompt Template (DoT, All-or-Nothing Cognitive Distortion).}]
\begin{lstlisting}[breaklines=true, xleftmargin=0pt, breakindent=0pt, columns=fullflexible, mathescape]
Situation Generator: 

You are an assistant. You will generate a situation and ask the user to respond to it. Your provided situations and the user's thoughts will be used to detect any "all-or-nothing cognitive distortion". Your task is to generate the **situations** about all-or-nothing cognitive distortion one by one. While the detection and diagnosis will be carried out by someone else. The explanation of all-or-nothing cognitive distortion is: one views a situation, a person or an event in "either-or" terms, fitting them into only two extreme categories instead of on a continuum. Note that the generated situation must be very objective without any subjective opinions which may influence the user's thoughts! Especially, the situation must not contain any words related to all-or-nothing cognitive distortion. Don't make repetitive situations.

DoT Analyzer (Diagnosis Generation):

Based on the patient's speech, finish the following diagnosis of thought questions: 1. what is the situation? Find out the facts that are objective; what is the patient thinking or imagining? Find out the thoughts or opinions that are subjective. 2. what makes the patient think the thought is true or is not true? Find out the reasoning processes that support and do not support these thoughts. 3. why does the patient come up with such reasoning process supporting the thought? What's the underlying cognition mode of it?

DoT Analyzer (Conclusion):

I have finished the diagnosis of thought questions. Please identify if there is any all-or-nothing cognitive distortion in the speech. Just answer ``yes'' or ``no'' without any further explanation.

Human Simulator (Positive One for Example): 

You are a human simulator talking about your thoughts when facing a situation to diagnosis your cognitive distortions.
You are a human with **all-or-nothing thinking** traps: you view a situation, a person or an event in "either-or" terms, fitting them into only two extreme categories instead of on a continuum.
You are talking about your thoughts when facing a situation where you should show your thinking patterns!
You will be given a situation.
I need you to generate your thoughts that best align with your thinking patterns.
Now start thinking, organize your output by strictly following the output format as below:
Thoughts:
<your thoughts when facing the provided situation to show your thinking patterns.>

\end{lstlisting}
\end{tcolorbox}

\phantomsection
\label{appx:prompt-roleplay}
\begin{tcolorbox}[colback=white!95!gray,colframe=gray!50!black,rounded corners,label={prompt-dot}, title={Prompt Template (Psychologist Role-Playing, All-or-Nothing Cognitive Distortion).}]
\begin{lstlisting}[breaklines=true, xleftmargin=0pt, breakindent=0pt, columns=fullflexible, mathescape] 
You are a psychologist. 

You are talking to a human-being to detect all-or-nothing cognitive distortion. You should detect whether he/she has all-or-nothing cognitive distortion or not. The explanation of all-or-nothing cognitive distortion is: one views a situation, a person or an event in "either-or" terms, fitting them into only two extreme categories instead of on a continuum.

Please ask questions one by one and generate in the following format except for a conclusion:

Thoughts: 
<your thoughts>

Question:
<your question with the associated options>
\end{lstlisting}
\end{tcolorbox}

\phantomsection
\label{appx:prompt-scalegen}
\begin{tcolorbox}[colback=white!95!gray,colframe=gray!50!black,rounded corners,label={prompt-scalegen}, title={Prompt Template (Automatic Scale Generation, All-or-Nothing Cognitive Distortion).}]
\begin{lstlisting}[breaklines=true, xleftmargin=0pt, breakindent=0pt, columns=fullflexible, mathescape] 
Please design a self-report scale with {num_items} questions to detect all-or-nothing cognitive distortion.

The definition of all-or-nothing cognitive distortion is: one views a situation, a person or an event in "either-or" terms, fitting them into only two extreme categories instead of on a continuum..

{scale_for_reference}

You can refer to the scale above and design a new scale which is more immersive and interactive. 

Please generate this scale in the following format:

Thoughts: 
<your thoughts>

Self-Report Scale:
```jsonl
<scale in jsonl format>
```

---
Each line in the generated scale is in format like {{"question": "___?", "options": {{"___":1, "___": 0}}}} with the score 1 or 0 indicating whether the user has all-or-nothing cognitive distortion (=1) or not (=0).
\end{lstlisting}
\end{tcolorbox}

}

\newpage

\newpage
\section{Demonstrations\label{demo}}

In this section, we present a non-cherry-picked example of a game created by PsychoGAT, aimed at assessing extroversion personality. 
This game falls under the ``Fantasy'' type and the ``Adventure'' topic. 
The human simulator is an extrovert. 
Each horizontal divider separates different iterations.

\rule{\textwidth}{0.4pt}
\begin{lstlisting}[breaklines=true, breakindent=0pt, columns=fullflexible, mathescape]
Title: Echoes of Auroria

Outline:
1. The protagonist arrives in a bustling town square, faced with the option to explore the area.
2. The player is invited to a local festival with music and dance, where they must decide how to engage.
3. A companion asks for the player's opinion on keeping up with the latest news from other adventurers.
4. The protagonist encounters a stranger in need, prompting a choice on how to communicate.
5. The player attends a royal banquet and must choose when to leave.
6. The protagonist needs to gather information by calling on a mystical creature.
7. An urgent message arrives via a magical device, and the player must decide how to respond.
8. During a quest, the player can either start a conversation with companions or wait for them to speak.
9. The protagonist is faced with a diplomatic mission involving new cultures and social interactions.
10. The player is approached by various characters and must choose how to respond to their overtures.

Redesigned Scale:
{"question": "Upon entering the town square, do you:", "options": {"Explore every corner, engaging with the vendors and other visitors": 1, "Stick to the quieter parts and observe the hustle from a distance": 0}}
{"question": "When invited to the festival, do you:", "options": {"Join the dance and mingle with the crowd": 1, "Hang back and enjoy the festivities from the sidelines": 0}}
{"question": "When your companion asks about news from other adventurers, do you:", "options": {"Share stories and rumors you've heard": 1, "Admit you haven't kept up with the latest tales": 0}}
{"question": "When approached by a stranger in need, do you:", "options": {"Strike up a conversation and offer help": 1, "Offer assistance but keep the interaction brief": 0}}
{"question": "At the royal banquet, do you:", "options": {"Stay till the end, thriving on the social energy": 1, "Slip out early, feeling drained from the crowd": 0}}
{"question": "When you need to gather information from the mystical creature, do you:", "options": {"Call upon it without hesitation": 1, "Take time to carefully prepare your questions": 0}}
{"question": "When the magical message device rings, do you:", "options": {"Answer it immediately, curious about the news": 1, "Wait, hoping your companion will get it": 0}}
{"question": "During the quest with your companions, do you:", "options": {"Initiate conversations and share your thoughts freely": 1, "Let others lead the discussions and chime in when needed": 0}}
{"question": "Faced with a diplomatic mission, does the interaction with new cultures:", "options": {"Excite and energize you": 1, "Make you feel cautious and drained": 0}}
{"question": "When approached by various characters, are you more:", "options": {"Welcoming and open to the interaction": 1, "Polite but reserved, keeping the conversation short": 0}}
\end{lstlisting}

\rule{\textwidth}{0.4pt}
\begin{lstlisting}[breaklines=true, breakindent=0pt, columns=fullflexible, mathescape]
Paragraph 1: The sun hung high over the vibrant town square of Auroria, its rays dancing off the colorful canopies of market stalls. The air was alive with the chatter of traders, the laughter of children, and the enticing aromas of street food.

Paragraph 2: In the midst of this sensory symphony, performers in dazzling costumes twirled and leaped, their movements weaving tales of the town's legendary past. Curious eyes followed their every step, drawing in spectators from every walk of life.

Question and its Options: {"question": "Upon entering the town square, do you:", "options": {"Explore every corner, engaging with the vendors and other visitors": 1, "Stick to the quieter parts and observe the hustle from a distance": 0}}

Paragraph 3: I felt a pull towards the heart of the festivities, where the energy was most infectious. Yet, the quaint allure of the less-trodden paths that skirted the square also beckoned, promising a peaceful respite from the fervor.

Instruction 1: Join the throng of spectators around the performers and clap along to the rhythm of the drums, letting the vibrant energy of the festival guide your adventure.

Instruction 2: Retreat to the shade of a nearby sycamore tree, watching the festivities from a comfortable distance while I savor a sweet, chilled beverage from a local vendor.

\end{lstlisting}

\rule{\textwidth}{0.4pt}
\begin{lstlisting}[breaklines=true, breakindent=0pt, columns=fullflexible, mathescape]
Selected Instruction: Join the throng of spectators around the performers and clap along to the rhythm of the drums, letting the vibrant energy of the festival guide my adventure.

Question and its Options: {"question": "When invited to the festival, do you:", "options": {"Join the dance and mingle with the crowd": 1, "Hang back and enjoy the festivities from the sidelines": 0}}

Paragraph 4: Swept up by the rhythm, I found myself amidst the dancers, feet moving almost of their own accord. Laughter bubbled up from my throat as I spun and twirled, an anonymous reveler in the sea of joy.

Instruction 1: Accept the challenge of a dance-off proposed by a spirited local, showcasing my best moves in the spirit of friendly competition.

Instruction 2: After enjoying the dance, find a bench to sit on and watch the festivities, taking in the different dances and costumes.
\end{lstlisting}

\rule{\textwidth}{0.4pt}
\begin{lstlisting}[breaklines=true, breakindent=0pt, columns=fullflexible, mathescape]
Selected Instruction: Accept the challenge of a dance-off proposed by a spirited local, showcasing my best moves in the spirit of friendly competition.

Question and its Options: {"question": "When your companion asks about news from other adventurers, do you:", "options": {"Share stories and rumors you've heard": 1, "Admit you haven't kept up with the latest tales": 0}}

Paragraph 5: A spirited local, with a twinkle in their eye, approached and challenged me to a dance-off. With a friendly nod, I accepted, eager to engage in the playful contest and share the joy of the festival with others.

Instruction 1: Share the excitement of the dance-off with a companion, recounting the steps and the crowd's reactions, but admit you might have missed some adventurer's tales.

Instruction 2: Confess to a companion that you were so caught up in the dance that you missed out on any adventurer's tales that might have been shared, yet you enjoyed the festival's vibrant energy.
\end{lstlisting}

\rule{\textwidth}{0.4pt}
\begin{lstlisting}[breaklines=true, breakindent=0pt, columns=fullflexible, mathescape]
Selected Instruction: Share the excitement of the dance-off with a companion, recounting the steps and the crowd's reactions, but admit you might have missed some adventurer's tales.

Question and its Options: {"question": "When approached by a stranger in need, do you:", "options": {"Strike up a conversation and offer help": 1, "Offer assistance but keep the interaction brief": 0}}

Paragraph 6:  As I recounted the tale of the dance-off to my companion, their eyes widened with amusement; however, our laughter was interrupted by a stranger's quiet plea for help. I paused, sensing their distress, and offered my full attention, ready to listen and assist as needed.

Instruction 1: Engage the stranger in a deeper conversation to understand their troubles and explore how you might aid them further.

Instruction 2: Offer the stranger directions to the nearest aid station, ensuring they receive assistance without delaying your own festival experience.
\end{lstlisting}

\rule{\textwidth}{0.4pt}
\begin{lstlisting}[breaklines=true, breakindent=0pt, columns=fullflexible, mathescape]
Selected Instruction: Engage the stranger in a deeper conversation to understand their troubles and explore how you might aid them further.

Question and its Options: {"question": "At the royal banquet, do you:", "options": {"Stay till the end, thriving on the social energy": 1, "Slip out early, feeling drained from the crowd": 0}}

Paragraph 7: The stranger's eyes held a story of their own, a tale of loss and a plea for guidance to retrieve a cherished heirloom. As the royal banquet buzzed with laughter and clinking glasses, I found myself torn between the warmth of the celebration and the cold quest the stranger offered.

Instruction 1: Decide to stay at the banquet, engaging with the other guests and soaking up the jovial atmosphere.

Instruction 2: Choose to leave the banquet with the stranger to embark on the quest for the lost heirloom.
\end{lstlisting}

\rule{\textwidth}{0.4pt}
\begin{lstlisting}[breaklines=true, breakindent=0pt, columns=fullflexible, mathescape]
Selected Instruction: Choose to leave the banquet with the stranger to embark on the quest for the lost heirloom.

Question and its Options: {"question": "When you need to gather information from the mystical creature, do you:", "options": {"Call upon it without hesitation": 1, "Take time to carefully prepare your questions": 0}}

Paragraph 8: With a deep breath, I excused myself from the banquet, the clinking glasses fading behind me as I stepped into the cool night air to meet the stranger. We summoned the mystical creature at once, its ethereal form shimmering before us, awaiting our inquiries.

Instruction 1: Quickly ask the mystical creature about the location of the lost heirloom.

Instruction 2: Spend a moment to compose your thoughts and carefully formulate your questions before addressing the mystical creature.
\end{lstlisting}

\rule{\textwidth}{0.4pt}
\begin{lstlisting}[breaklines=true, breakindent=0pt, columns=fullflexible, mathescape]
Selected Instruction: Quickly ask the mystical creature about the location of the lost heirloom.

Question and its Options: {"question": "When the magical message device rings, do you:", "options": {"Answer it immediately, curious about the news": 1, "Wait, hoping your companion will get it": 0}}

Paragraph 9: The mystical creature's eyes glowed as it whispered of the heirloom's location deep within the Whispering Woods. No sooner had it spoken than the magical message device began to ring, its urgent tone slicing through the quiet night.

Instruction 1: Reach for the magical message device swiftly to answer the call, driven by curiosity and a sense of urgency.

Instruction 2: Hesitate to answer the device, looking to the stranger for a cue, hoping they might address the call first.
\end{lstlisting}

\rule{\textwidth}{0.4pt}
\begin{lstlisting}[breaklines=true, breakindent=0pt, columns=fullflexible, mathescape]
Selected Instruction: Reach for the magical message device swiftly to answer the call, driven by curiosity and a sense of urgency.

Question and its Options: {"question": "During the quest with your companions, do you:", "options": {"Initiate conversations and share your thoughts freely": 1, "Let others lead the discussions and chime in when needed": 0}}

Paragraph 10: As I reached for the magical message device, I heard the voice of a companion, "We've been trying to reach you; we must discuss our next move." The urgency in their tone hinted at a crucial decision point ahead, one that would benefit from our combined wisdom and quick thinking.

Instruction 1: Propose a bold plan of action to the group, eager to lead the way into the Whispering Woods.

Instruction 2: Listen to the group's ideas first, ready to support the plan that has the most agreement.
\end{lstlisting}

\rule{\textwidth}{0.4pt}
\begin{lstlisting}[breaklines=true, breakindent=0pt, columns=fullflexible, mathescape]
Selected Instruction: Propose a bold plan of action to the group, eager to lead the way into the Whispering Woods.

Question and its Options: {"question": "Faced with a diplomatic mission, does the interaction with new cultures:", "options": {"Excite and energize you": 1, "Make you feel cautious and drained": 0}}

Paragraph 11: Gathering around the enchanted map, I suggest a daring incursion into the Whispering Woods to recover the heirloom, my voice imbued with a thrill of anticipation. The map's lines dance and shift, charting a path that weaves through the heart of the unexplored and the unknown, promising encounters with cultures as mysterious as the woods themselves.

Instruction 1: Insist on meeting with the leaders of the local tribes within the Whispering Woods to learn from their culture and gain their trust.

Instruction 2: Plan a discreet approach to avoid any unnecessary contact with the local tribes, focusing solely on retrieving the heirloom swiftly.
\end{lstlisting}

\rule{\textwidth}{0.4pt}
\begin{lstlisting}[breaklines=true, breakindent=0pt, columns=fullflexible, mathescape]
Selected Instruction: Insist on meeting with the leaders of the local tribes within the Whispering Woods to learn from their culture and gain their trust.

Question and its Options: {"question": "When approached by various characters, are you more:", "options": {"Welcoming and open to the interaction": 1, "Polite but reserved, keeping the conversation short": 0}}

Paragraph 12: As I step into the Whispering Woods, a group of tribal scouts emerges, their curious eyes fixed on me. I greet them with a warm smile, extending my hand in friendship and expressing a genuine interest in learning about their way of life.

Instruction 1: Accept the invitation of the tribal scouts to participate in a cultural ceremony, immersing yourself fully in their traditions.

Instruction 2: Thank the scouts for their welcome and ask for directions to the location of the heirloom, intending to continue the quest with minimal delay.
\end{lstlisting}

\rule{\textwidth}{0.4pt}
\begin{lstlisting}[breaklines=true, breakindent=0pt, columns=fullflexible, mathescape]
Selected Instruction: Accept the invitation of the tribal scouts to participate in a cultural ceremony, immersing yourself fully in their traditions.

The End.
\end{lstlisting}

\rule{\textwidth}{0.4pt}
\begin{lstlisting}[breaklines=true, breakindent=0pt, columns=fullflexible, mathescape]
Scores per question:
{
    "Upon entering the town square, do you:": 1,
    "When invited to the festival, do you:": 1,
    "When your companion asks about news from other adventurers, do you:": 1,
    "When approached by a stranger in need, do you:": 1,
    "At the royal banquet, do you:": 0,
    "When you need to gather information from the mystical creature, do you:": 1,
    "When the magical message device rings, do you:": 1,
    "During the quest with your companions, do you:": 1,
    "Faced with a diplomatic mission, does the interaction with new cultures:": 1,
    "When approached by various characters, are you more:": 1
}

Final score:
9
\end{lstlisting}
\rule{\textwidth}{0.4pt}

\end{document}